\documentclass[letterpaper]{article} 
\usepackage{aaai23}  
\usepackage{times}  
\usepackage{helvet}  
\usepackage{courier}  
\usepackage[hyphens]{url}  
\usepackage{graphicx} 
\usepackage{natbib}  
\usepackage{caption} 
\usepackage{url}            
\usepackage{booktabs}       
\usepackage{amsfonts}       

\usepackage{microtype}      
\usepackage{xcolor}         
\usepackage{enumitem}
\usepackage[english]{babel}
\usepackage{amsthm}
\usepackage{wrapfig}
\usepackage{subfig}  
\usepackage[bookmarksnumbered=true]{hyperref} 

\hypersetup{
     colorlinks = true,
     linkcolor = black,
     anchorcolor = black,
     citecolor = black,
     filecolor = black,
     urlcolor = purple
     }

\newtheorem{definition}{Definition}

\usepackage{algorithm}
\usepackage{algorithmic}
\newcommand{\etal}{\textit{et al}. }
\newcommand{\ie}{\textit{i}.\textit{e}., }
\newcommand{\eg}{\textit{e}.\textit{g}., } 
\usepackage{newfloat}
\usepackage{listings}
\DeclareCaptionStyle{ruled}{labelfont=normalfont,labelsep=colon,strut=off} 
\floatstyle{ruled}
\newfloat{listing}{tb}{lst}{}
\floatname{listing}{Listing}
\title{Exploring Temporal Information Dynamics in Spiking Neural Networks}

\author{
    Youngeun Kim\textsuperscript{\rm 1}, Yuhang Li\textsuperscript{\rm 1}, Hyoungseob Park\textsuperscript{\rm 1}, Yeshwanth Venkatesha\textsuperscript{\rm 1}, \\ Anna Hambitzer\textsuperscript{\rm 2}, Priyadarshini Panda\textsuperscript{\rm 1} 
}
\affiliations{

\textsuperscript{\rm 1} Department of Electrical Engineering, Yale University, New Haven, CT, USA\\
\textsuperscript{\rm 2} Technology Innovation Institute, Abu Dhabi, UAE \\
{\{youngeun.kim, yuhang.li, hyoungseob.park, yeshwanth.venkatesha, priya.panda\}@yale.edu},  Anna.Hambitzer@tii.ae
}

\usepackage{bibentry}

\begin{document}

\maketitle

\begin{abstract}
Most existing Spiking Neural Network (SNN) works state that SNNs may utilize temporal information dynamics of spikes. However, an explicit analysis of temporal information dynamics is still missing. In this paper, we ask several important questions for providing a fundamental understanding of SNNs: \textit{What are temporal information dynamics inside SNNs? How can we measure the temporal information dynamics? How do the temporal information dynamics affect the overall learning performance?} To answer these questions, we estimate the Fisher Information of the weights to measure the distribution of temporal information during training in an empirical manner.
Surprisingly, as training goes on, Fisher information starts to concentrate in the early timesteps. After training, we observe that information becomes highly concentrated in earlier few timesteps, a phenomenon we refer to as \textit{temporal information concentration}. We observe that the temporal information concentration phenomenon is a common learning feature of SNNs by conducting extensive experiments on various configurations such as architecture, dataset, optimization strategy, time constant, and timesteps. Furthermore, to reveal how temporal information concentration affects the performance of SNNs, we design a loss function to change the trend of temporal information. We find that temporal information concentration is crucial to building a robust SNN but has little effect on classification accuracy.
Finally, we propose an efficient iterative pruning method based on our observation on temporal information concentration. 
Code is available at \href{https://github.com/Intelligent-Computing-Lab-Yale/Exploring-Temporal-Information-Dynamics-in-Spiking-Neural-Networks}{https://github.com/Intelligent-Computing-Lab-Yale/Exploring-Temporal-Information-Dynamics-in-Spiking-Neural-Networks}. 
\end{abstract}

\section{Introduction}
Within the last decade, Spiking Neural Networks (SNNs) have received huge attention as a low-power alternative to Artificial Neural Networks (ANNs) \cite{roy2019towards,christensen20222022,lobo2020spiking,wang2020supervised}. SNNs process visual information in an event-driven manner using sparse binary spikes over multiple timesteps that make them an attractive option for low-power neuromorphic hardware implementation \cite{akopyan2015truenorth,davies2018loihi,furber2014spinnaker}. 
Recent years have witnessed a surge in SNN algorithmic works that aim to improve accuracy of SNNs on standard image recognition tasks while maintaining higher efficiency than ANNs \cite{wu2019direct,comsa2020temporal,mostafa2017supervised,li2021free,deng2022temporal}.

Although most of the existing works on SNN assert that they might improve (or leverage) the temporal dynamics of spikes  \cite{wu2018spatio,fang2021incorporating,kim2020revisiting,masquelier2011timing,neftci2019surrogate}, yet, an explicit analysis of temporal information dynamics is still missing.
In this paper, we ask several important questions for understanding such fundamental characteristics of SNNs: 
\textit{What are temporal information dynamics inside SNNs? How can we measure the temporal information dynamics?  How do the temporal information dynamics affect the overall learning performance? }
Understanding the dynamics will enable us to apprehend the learning representations inside SNNs which may help to develop better temporal training algorithms, find new use-cases for SNNs for conventional AI (and possibly computational neuroscience) applications, and also explore new theoretical directions for SNN research.

To this end, we present a first-of-its-kind study to understand temporal information dynamics in SNNs through the lens of Fisher information.
In this study, we select a ResNet model with Leak-Integrated-and-Fire spiking neuron as a baseline, which is widely used in classification task \cite{zheng2020going,li2021differentiable,fang2021deep}. 
We measure the Fisher information of an SNN at each timestep, where we find the temporal information distribution varies as training goes on.
Specifically, we find that information in an SNN shifts from latter timesteps to earlier timesteps as training progresses. We call this phenomenon as \textit{temporal information concentration (TIC)}.
This is a novel observation, and therefore, we investigate whether the TIC phenomenon is a function of specific training variables such as architecture, dataset, and optimization strategy.
Through extensive experiments, we found that TIC is a common characteristic of SNNs during training.

Furthermore, we explore the impact of TIC on the performance of networks (\ie accuracy and robustness). 
To this end, we design a loss function that can manipulate the Fisher information at each timestep. 
The results show that TIC significantly affects the robustness of SNN against both adversarial perturbations and standard input noise (such as Gaussian and Blur). 
Unlike robustness, changes in temporal information due to TIC manipulation do not show any meaningful effect on classification accuracy. 
Finally, we propose an efficient iterative pruning strategy for SNNs using TIC, where we found the pruning performance is almost preserved with less number of timesteps.

In summary, our key contributions are as follows: (1) For the first time, we quantitatively analyze the temporal dynamics of SNNs. (2) Using Fisher information, we find that temporal information concentration (TIC) is a general trend in SNNs during training.  (3) By designing a loss forcing SNNs to have a different trend of TIC, we find that TIC plays a crucial role in imparting robustness to SNNs. 
(4) Finally, we apply the TIC observation to propose an efficient iterative pruning method for SNNs.

\section{Modeling Spiking Neural Networks }
\label{section:modelSNN}

In this section, we {briefly introduce} the neuron type and input encoding technique used in our analysis. 
We use Spiking Neural Networks (SNNs) with discretized Leaky Integrate-and-Fire (LIF) neurons \cite{roy2019towards, wu2019direct, fang2021incorporating,kim2022exploring} using simulation step $dt = 1$, which can be formulated by
\begin{equation}
    U_l^t = (1 - \frac{1}{\tau})  U_l^{t-1} + \frac{1}{\tau}  W_{l}O^t_{l-1}, 
    \label{eq:LIF}
\end{equation}
where  $U^t_l$ denotes membrane potential at time-step $t$ for layer $l$, and $O^t_{l-1}$ stands for the spike output from the previous layer. Also, $W_l$ represents weight connections at layer $l$, and $\tau$ is a time constant for decaying the membrane potential. Note, capital letters (\eg $U^t_l$ and  $O^t_{l-1}$) represent matrices.
The neuron generates a spike output if its membrane potential exceeds a firing threshold. Then, the membrane potential is reset to zero after firing.
For training weight parameters, we use spatio-temporal back-propagation (STBP), which accumulates the gradients over all timesteps \cite{wu2018spatio,neftci2019surrogate}.
We can formulate the gradients at the layer $l$ by chain rule, given by
\begin{equation}
      \Delta W_l =\frac{\partial L}{\partial W_l} =
 \sum_{t}(\frac{\partial L}{\partial O_l^t}\frac{\partial O_l^t}{\partial U_l^t} + \frac{\partial L}{\partial U_l^{t+1}}  \frac{\partial U_l^{t+1}}{\partial U_l^{t}})
 \frac{\partial U_l^t}{\partial W_l}.
\label{eq:delta_W}
\end{equation}
While computing backward gradients,  we use an approximate piece-wise linear function gradient $\frac{1}{\pi}arctan(\pi x) + \frac{1}{2}$ in order to address non-differentiability of LIF neurons \cite{fang2021incorporating}.
Overall, according to gradient descent, the network parameters are updated as $W_l = W_l - \eta \Delta W_l$, where $\eta$ represents learning rate. 

In this work, we focus on static image recognition where the majority of prior SNN works have focused so far \cite{roy2019towards}. 
We choose to generate spikes in an end-to-end manner, \ie directly encode the images in the first layer, due to its simplicity, flexibility, and high performance on large-scale datasets \cite{wu2019direct,zheng2020going,zhang2020temporal,fang2021incorporating,lee2020enabling,li2021free,kim2022neural}.
A given image is shown for $T$ timesteps $\{1,2,\dots,T\}$ to an SNN, and the final prediction is computed by accumulating the spikes at the output layer across $T$ steps.

\begin{figure*}[t]
\begin{center}
\def\arraystretch{0.5}
\hspace{-4mm}
\includegraphics[width=0.95\linewidth]{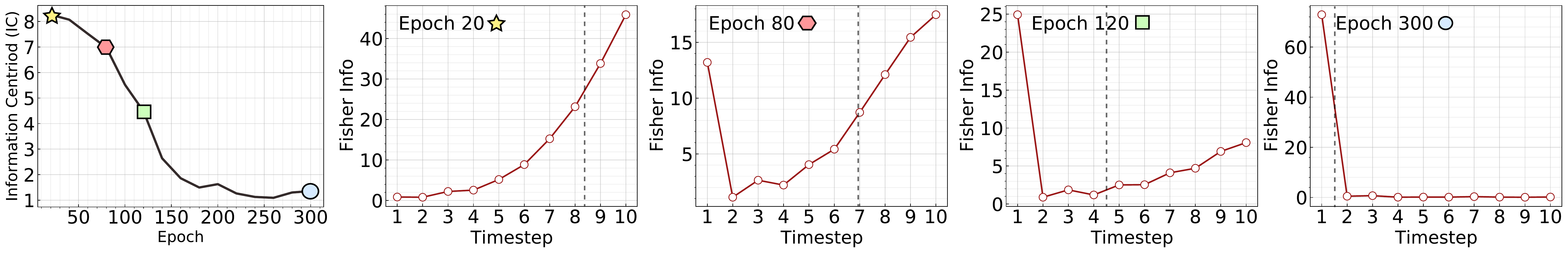}
\\
 \vspace{-3.5mm}
\end{center}
\caption{ 
Illustration of temporal dynamics across training epochs.
Starting from the left panel, we present information centroid (Eq. \ref{eq:information_centriod}) across epochs.
We also report Fisher information in temporal dimension (Eq. \ref{eq:final_information}) at four different epochs (marked with different colors).
As training goes on, the early timesteps obtain more information, while the information decreases in the latter timesteps.
Accordingly, the information centroid (vertical dotted line) moves towards early timesteps as training progresses.
}
 \vspace{-2mm}
\label{fig:method:fisher_train_base}
\end{figure*}

\section{Fisher Information Analysis in Time Dimension}
The Fisher Information Matrix (FIM) quantifies the amount of information inside a model obtained from a given data, when the model parameters are perturbed \cite{fisher1925theory}.
If weight perturbation brings small (or large) change in the model's prediction, we can say that the model contains small (or large) information with respect to the corresponding data.
 FIM can be interpreted as the second-order gradient of KL divergence between the original model and the weight-perturbed model \cite{achille2018critical}.
 Mathematically, given a network's approximate posterior distribution $f_{\theta}(y|x)$ with weight parameters $\theta$, input image $x$ sampled from data distribution $D$, output variable $y$, FIM can be formulated as follows:
\begin{equation}
    M = \mathbb{E}_{x \sim D}  \mathbb{E}_{y \sim f_{\theta}(y|x)} [ \nabla_{\theta} log f_{\theta}(y|x) \nabla_{\theta} log f_{\theta}(y|x)^T].
    \label{eq:fisher_mat}
\end{equation}
 Existing works have utilized FIM to explore the characteristic of the loss landscape \cite{keskar2016large,liang2019fisher,soen2021variance},  model capacity \cite{ly2017tutorial}, and model training dynamics \cite{achille2018critical}.
 
Different from the conventional ANN model, SNN predicts class probabilities by accumulating a given input through multiple timesteps. 
Our objective is to analyze the information dynamics of the model across time.
Therefore, we introduce a metric to measure the amount of accumulated FIM in SNN from timestep $1$ to timestep $t$:
\begin{equation}
    M_{t} \hspace{1mm} {=}  \hspace{1mm} \mathbb{E}_{x \sim D}  \mathbb{E}_{y \sim f_{\theta}(y|x_{i\le t})} [ \nabla_{\theta} log f_{\theta}(y|x_{i\le t}) \nabla_{\theta} log f_{\theta}(y|x_{i\le t})^T],
    \label{eq:temporal_fisher_mat}
\end{equation}
where, $i \in \{1,2,..., T\}$ is a positive integer that represents the index of timestep.
Note that it is difficult to measure the amount of information in one timestep independently since the posterior distribution of SNNs is based on the information from all previous timesteps with LIF neurons.
One major problem of FIM on deep neural networks is the size of the matrix, which is too large to compute completely. To address this, following previous works \cite{achille2018critical,kirkpatrick2017overcoming}, we use the trace of FIM to measure the accumulated information $I_t$ stored in weight parameters from timestep $1$ to $t$: 
\begin{equation} 
        I_t = Tr(M_t) 
        = \mathbb{E}_{x \sim D}  \mathbb{E}_{y \sim f_{\theta}(y|x_{i\le t})}[ \| \nabla_{\theta} log f_{\theta}(y|x_{i\le t}) \|^2 ].
    \label{eq:Trace_I}
\end{equation}
Given $N$ training samples, the expectation in Eq. \ref{eq:Trace_I} can be replaced by the empirical mean across observed data \cite{amari2000adaptive,martens2015optimizing,karakida2019universal}: 
\begin{equation}
    I_{t}  = \frac{1}{N} \ \sum_{n=1}^{N} \| \nabla_{\theta} log f_{\theta}(y|x^{n}_{i\le t}) \|^2.
    \label{eq:final_information}
\end{equation}
Across all experiments, we use Eq. \ref{eq:final_information} for quantifying the amount of Fisher information (or sometimes we use \textit{information} interchangeably) across different timesteps.

In ANN literature, the empirical FIM trace (Eq. \ref{eq:final_information}) can be interpreted as a measure for the importance of weight connections \cite{achille2018critical,kirkpatrick2017overcoming} with respect to training data.
For example, Kirkpatrick \etal \cite{kirkpatrick2017overcoming} address catastrophic forgetting in continual learning (\ie  a network learns different tasks sequentially) by maintaining connections having high FIM trace from the prior tasks.
Achille \etal \cite{achille2018critical} use empirical FIM trace in order to discover the important training epochs for standard deep neural networks, where they show that epochs having a higher FIM trace impact more on the accuracy.
Similarly, in our work, the SNN model having a high FIM trace at timestep $t$ represents that the weight connections inside the SNN contain important information for the given input until timestep $t$.

\textbf{Temporal information concentration in SNN.} 
To reveal the temporal information dynamics in SNNs, we first present temporal Fisher information of SNNs across training epochs.
In Fig. \ref{fig:method:fisher_train_base}, we visualize temporal Fisher information of a ResNet19 architecture on CIFAR10 dataset.
Interestingly, as training goes on, the amount of Fisher information in the latter timesteps steadily decreases while the early-time information increases.
In the final trained SNN model (epoch 300), Fisher information concentrates on the first few timesteps, and maintains a near-zero value till the end of timesteps. 
We call this phenomenon as \textit{temporal information concentration (TIC)} - an information shift from latter timesteps to the early timesteps as training progresses.

\begin{figure}[t]
  \centering
 \includegraphics[width=0.49\textwidth]{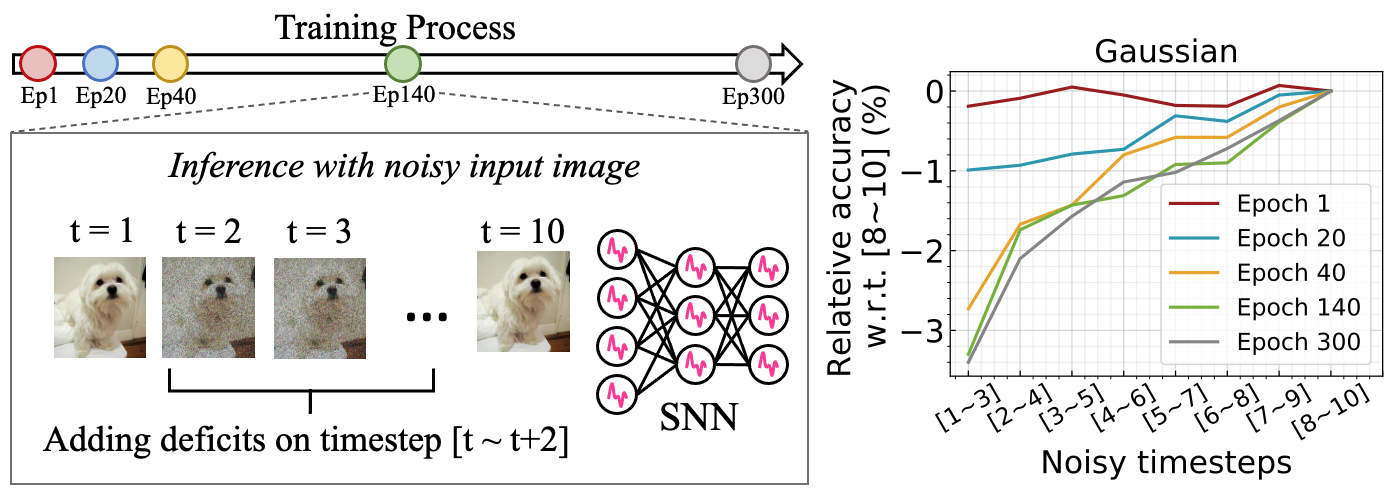}  \caption{ 
Additional experiments support TIC observation at inference.
  We use a ResNet19 architecture on the CIFAR10 dataset.
  We select five different SNN models during training (with clean data): The early training phase (epoch 1, 20), the intermediate phase (epoch 40, 140), and the final trained model (epoch 300). Then, we measure the test accuracy of the model at each point by adding deficits (\ie Gaussian noise) to the input image for time window $[t \sim t+2]$. We change $t$ from $1$ to $8$.
 In the right panels, we report the relative test accuracy w.r.t.  noise in $[8\sim10]$, \ie $Acc_{[t\sim t+2]} - Acc_{[8\sim10]}$.
We provide experimental details in Appendix.
 }
      \label{fig:intro:noise_train_infer}
    \vspace{-3mm}
\end{figure}

Further, to provide better visualization of the overall trend of information dynamics across epochs, we introduce a metric called \textit{Information Centroid} (IC). Given SNNs with $T$ timesteps, IC can be formulated by
\begin{equation}
    IC = \frac{\sum_{t \in \{1, ..., T\}} t I_t}{\sum_{t \in \{1, ..., T\}} I_t}.
    \label{eq:information_centriod}
\end{equation}
Thus, high IC means Fisher information ($I_t$) increases with timestep $t$ (\eg epoch 20 in Fig. \ref{fig:method:fisher_train_base}). In Fig. \ref{fig:method:fisher_train_base}, SNN at epoch 20 yields highest IC value which can be understood from its increasing $I_t$ trend across different timesteps. Small IC denotes that information concentrates in the early timesteps (\eg epoch 300 in Fig. \ref{fig:method:fisher_train_base} where $I_t$ is highest at $t=1$ and becomes nearly 0 for $t=2,..,10$).

Moreover, we found that TIC is closely related to performance degradation with deficits at inference.
As shown in Fig. \ref{fig:intro:noise_train_infer}, we select models from five different epochs, and add the deficits to the input image for a certain time window.
If the accuracy degrades significantly in a specific time window,  those timesteps are likely to convey critical information for prediction \cite{achille2018critical}.
At the beginning of the training phase (or early epochs), all timesteps show similar noise sensitivity. However, interestingly, as training goes on, early timesteps show higher noise sensitivity compared to the later timesteps.
The results support our observation on TIC, where the early timesteps contain important information as training evolves.

\begin{figure}[t]
\begin{center}
\def\arraystretch{0.5}
\begin{tabular}{@{}c@{}c@{}c@{}c@{}c@{}c}
\hspace{-4mm}
\includegraphics[width=0.45\linewidth]{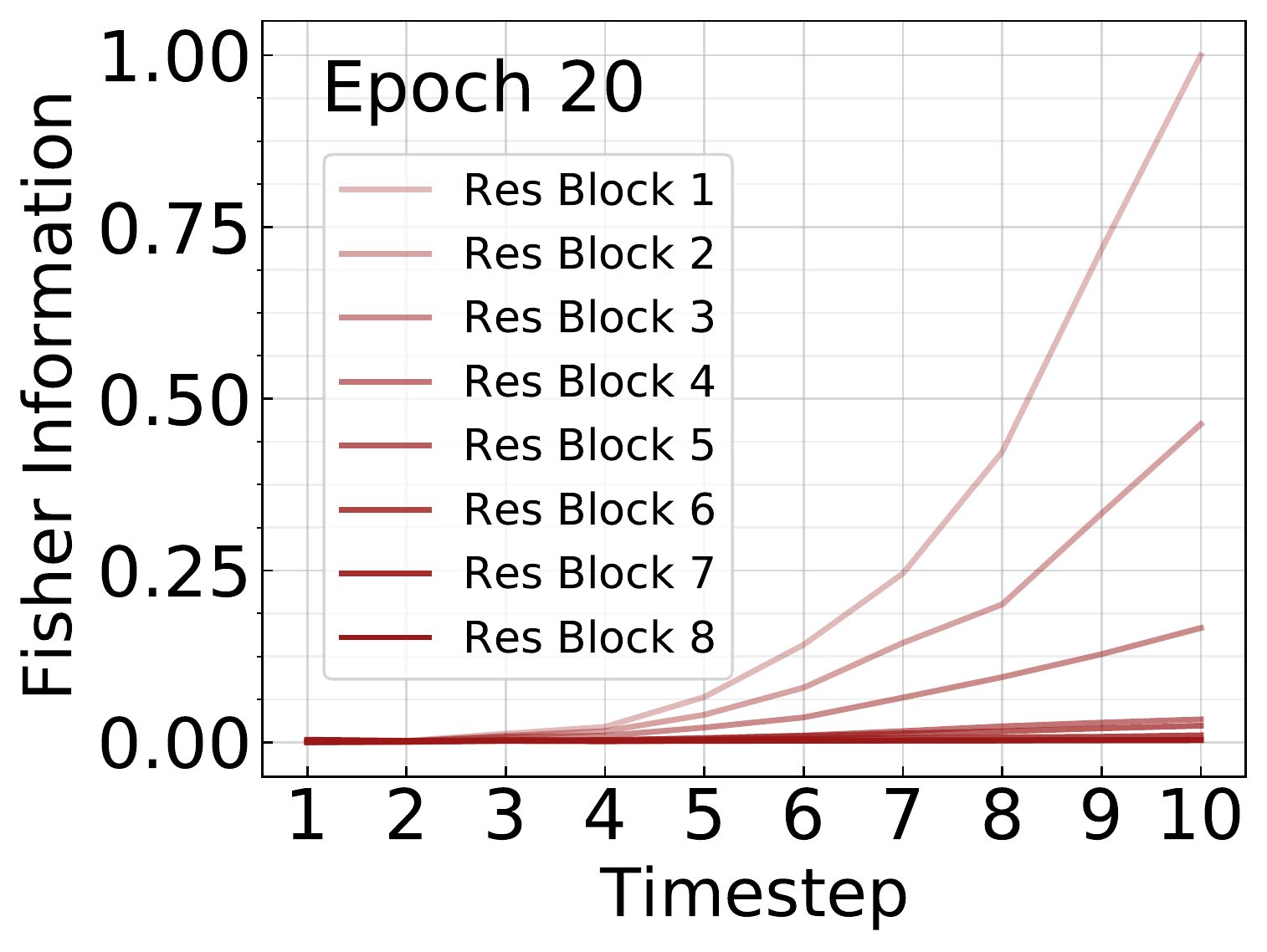} &
\includegraphics[width=0.45\linewidth]{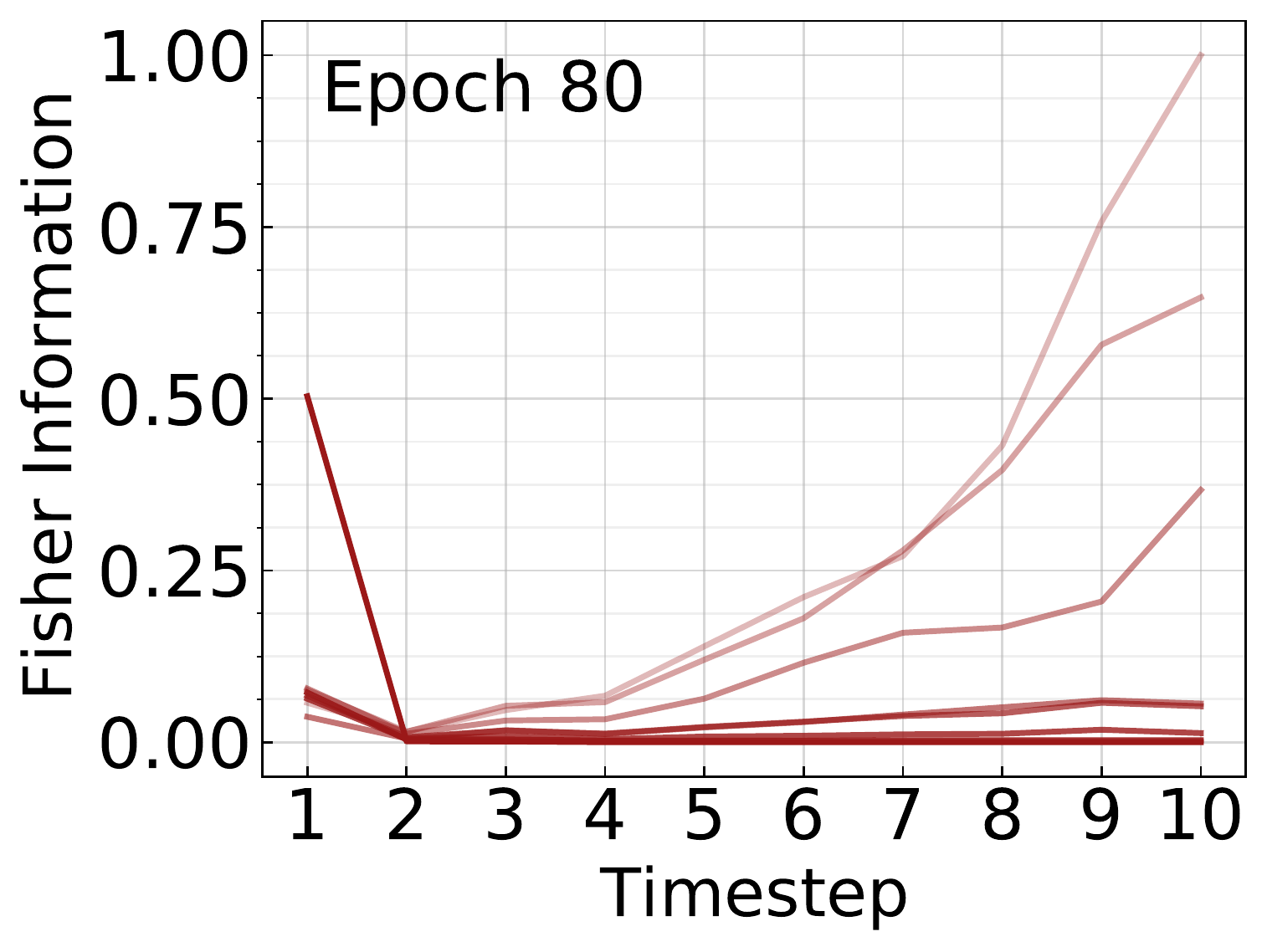} & \\
\includegraphics[width=0.45\linewidth]{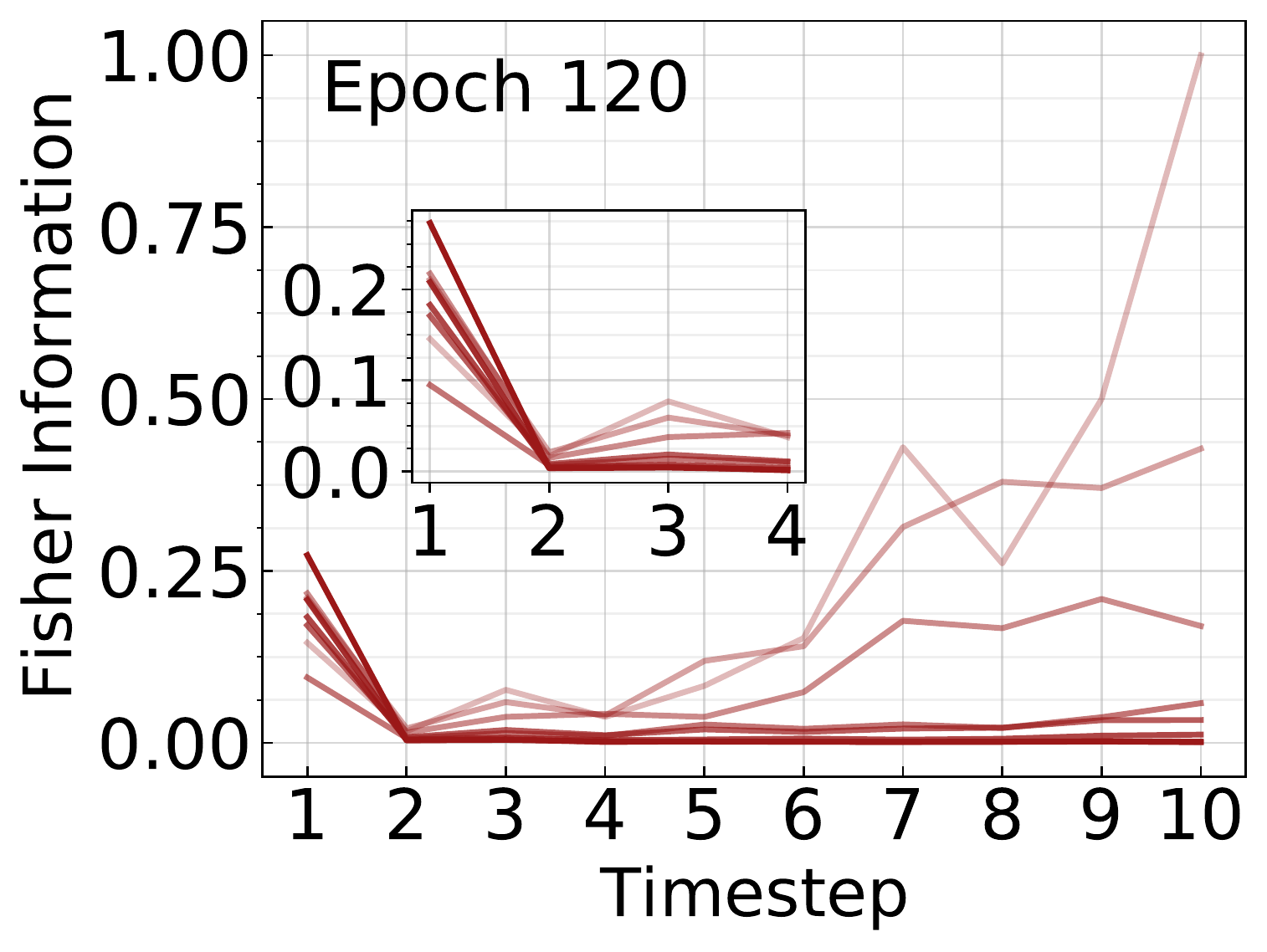} &
\includegraphics[width=0.45\linewidth]{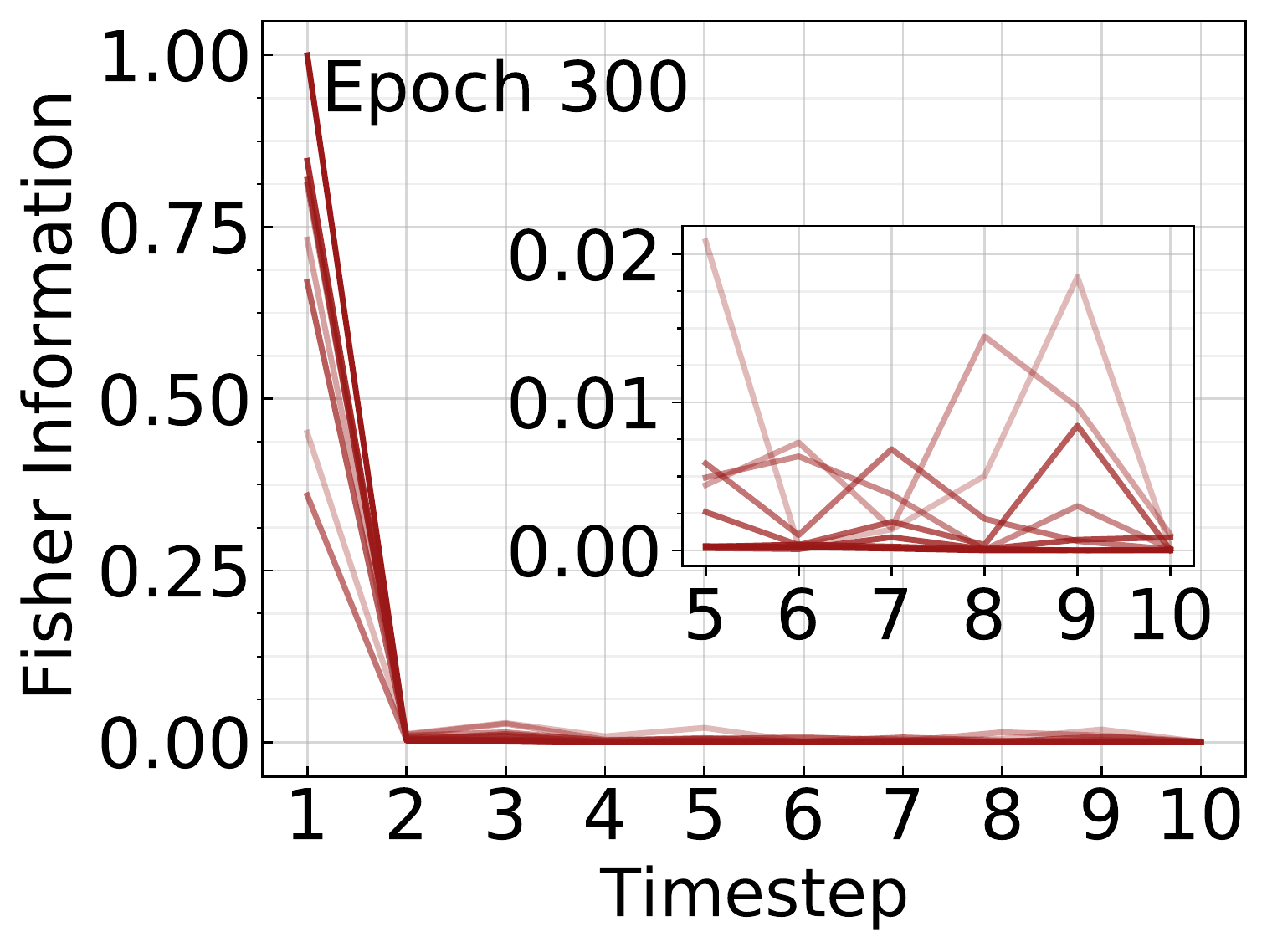} 

 \vspace{-2.5mm}
 \\
\end{tabular}
\end{center}
\caption{ 
   Normalized Fisher information contained in each residual block (in ResNet19 trained on CIFAR10) as a function of timestep. We visualize Fisher information at Epoch 20, 120 and 300. Here, we provide the normalized Fisher information for better visualization on relative Fisher information across all layers.
}
 \vspace{-4mm}
\label{fig:method:layer_prop}
\end{figure}

\textbf{Layer-wise analysis with Fisher Information.} 
Previous works \cite{kirkpatrick2017overcoming,achille2018critical} use Fisher information to measure the importance of layers (or weight connections).
If a layer contains high Fisher information, the layer has a high contribution to the prediction for the given data. Here, we conduct layer-wise Fisher information analysis to observe the importance of each layer across timesteps.
To this end, we measure Fisher information contained in each residual block of a ResNet19 SNN model (Fig. \ref{fig:method:layer_prop}).
The overall trend of temporal information dynamics in each layer follows the TIC trend.
Intriguingly, we find that, after the SNN model is trained for several epochs (\eg epoch 120 and 300), shallow and deep layers contain high Fisher information at the latter and early timesteps, respectively.
Such observation can be interpreted through the role of shallow and deep layers in feature representation. 
The prior ANN and SNN works  \cite{selvaraju2017grad,kim2021visual} have revealed that shallow and deep layers of a model contain low-level (\eg texture) and high-level (\eg semantic) features, respectively. 
Thus, high Fisher information in shallow/deep layers means that weight connections related to low/high-level features are important for prediction.
Accordingly, we conclude that the SNN model focuses on high-level features at the early timesteps, while low-level features are important at the latter timesteps for prediction.

\begin{figure}[t]
\begin{center}
\def\arraystretch{0.5}
\begin{tabular}{@{}c@{\hskip 0.028\linewidth}c@{\hskip 0.028\linewidth}c@{}c@{}c@{}c}
\hspace{-4mm}
\includegraphics[width=0.30\linewidth]{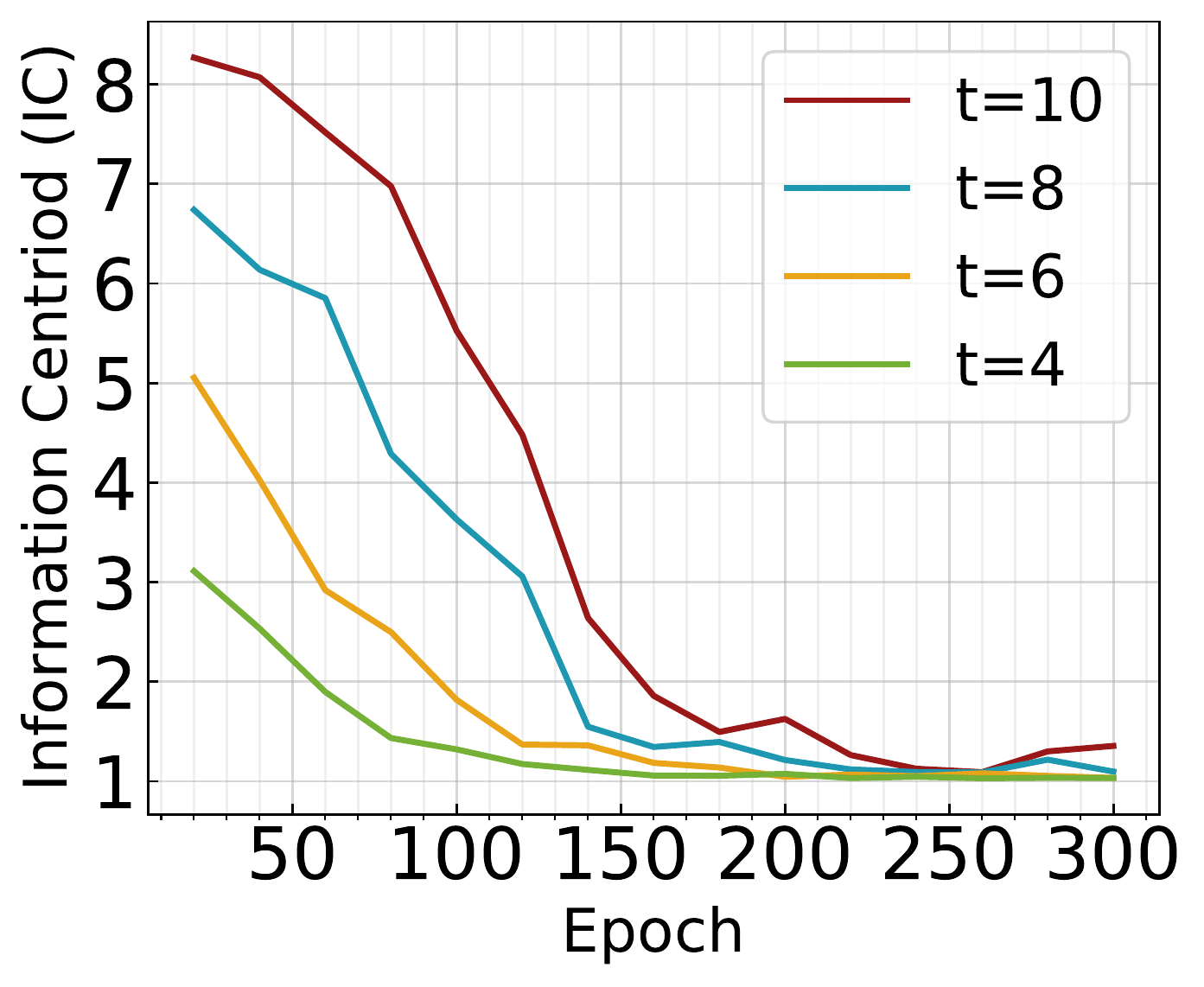} &\includegraphics[width=0.30\linewidth]{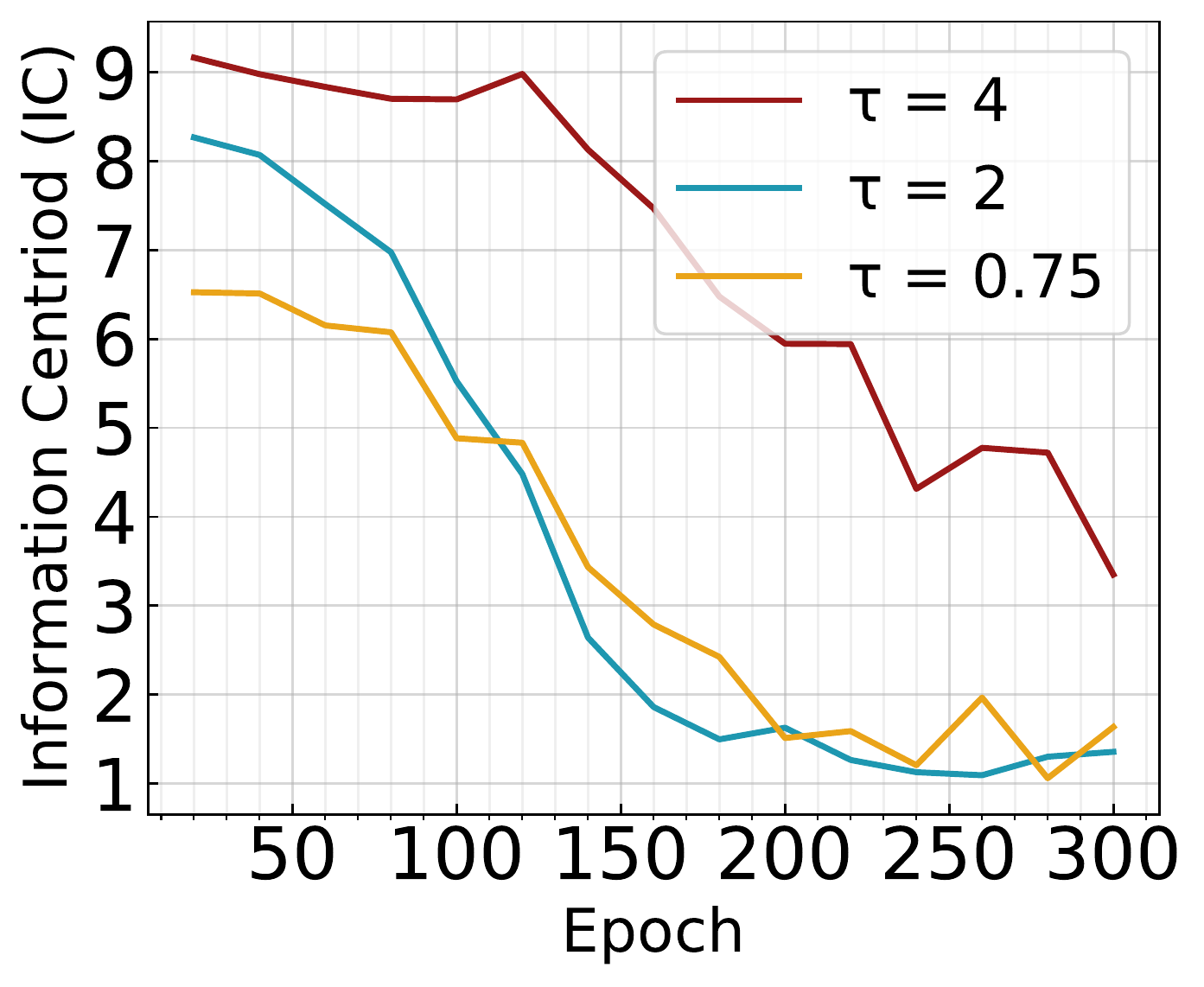} \hspace{-2mm}
\includegraphics[width=0.38\linewidth]{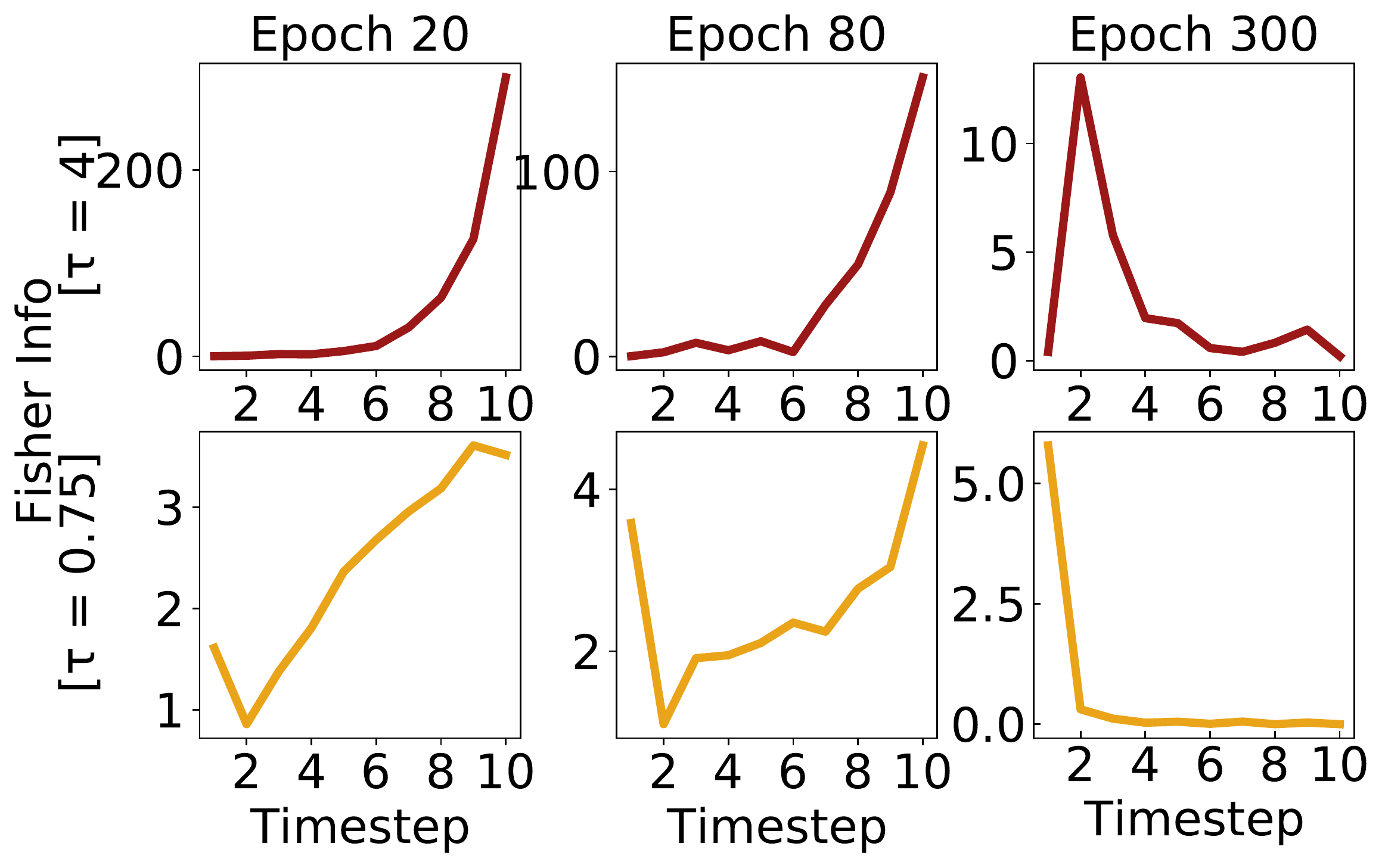} \\
{\hspace{1mm} (a) Timestep } & {\hspace{1mm} (b) Time constant}
\\
\includegraphics[width=0.30\linewidth]{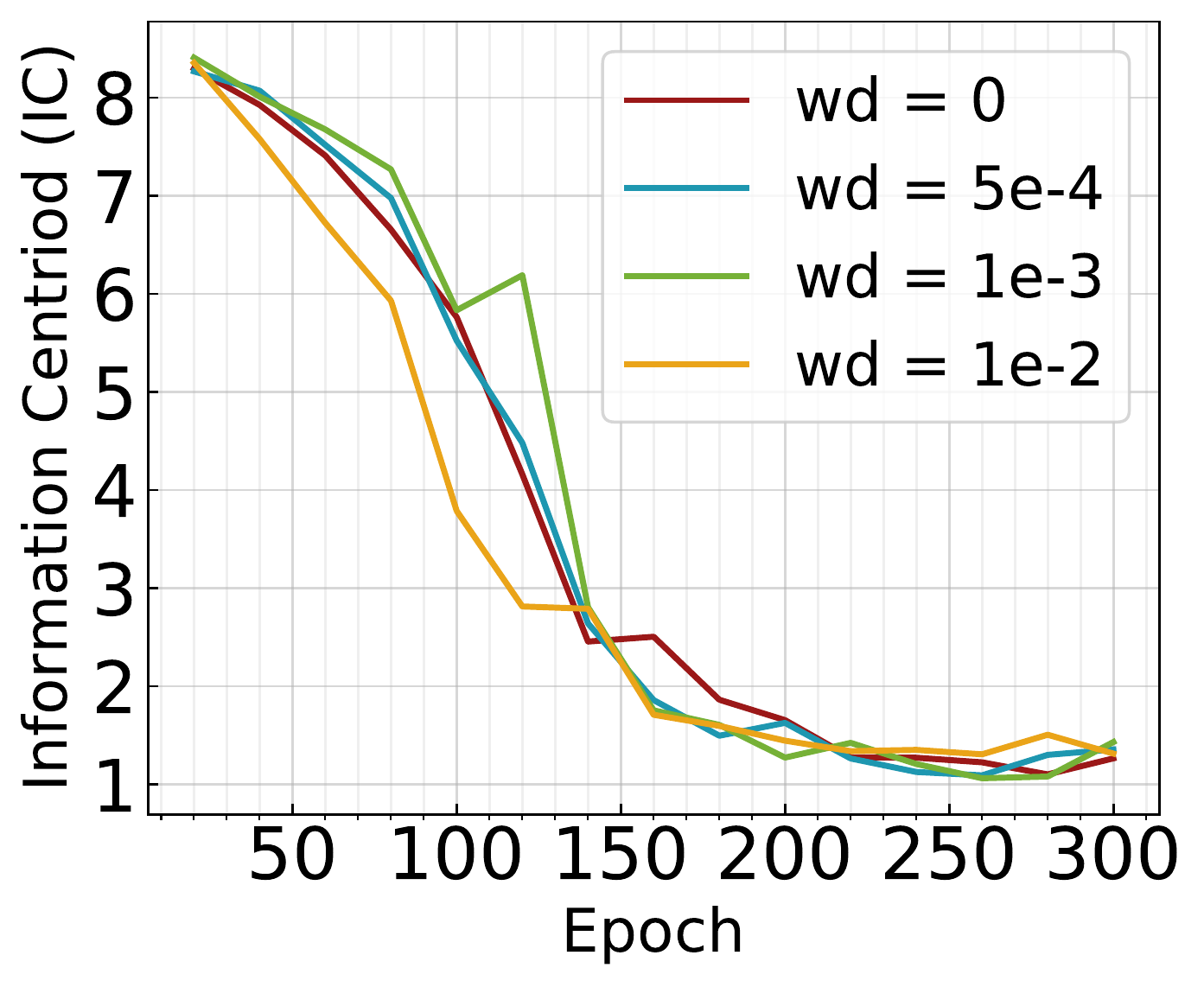}&
\includegraphics[width=0.30\linewidth]{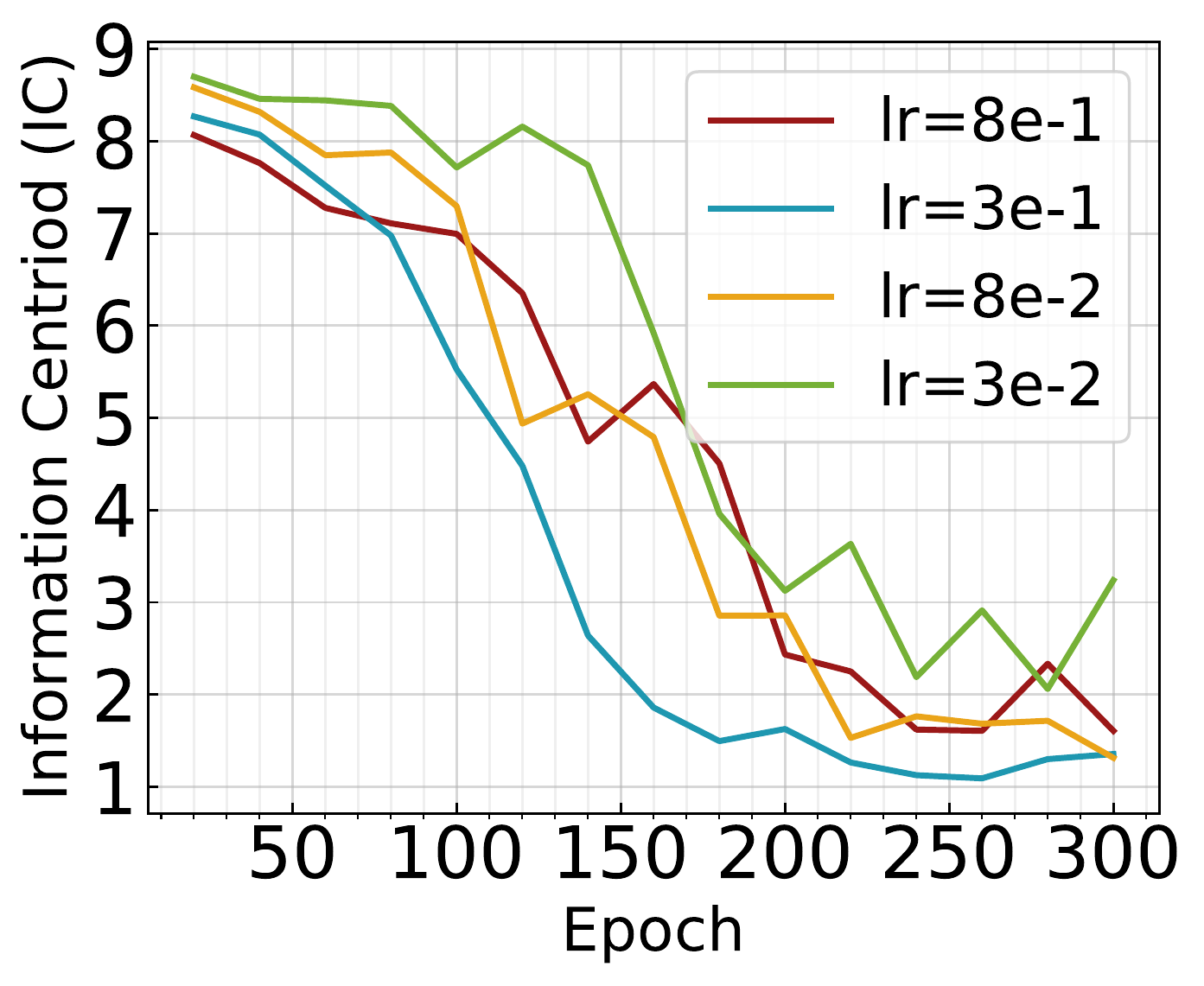}
 \vspace{1mm}\includegraphics[width=0.38\linewidth]{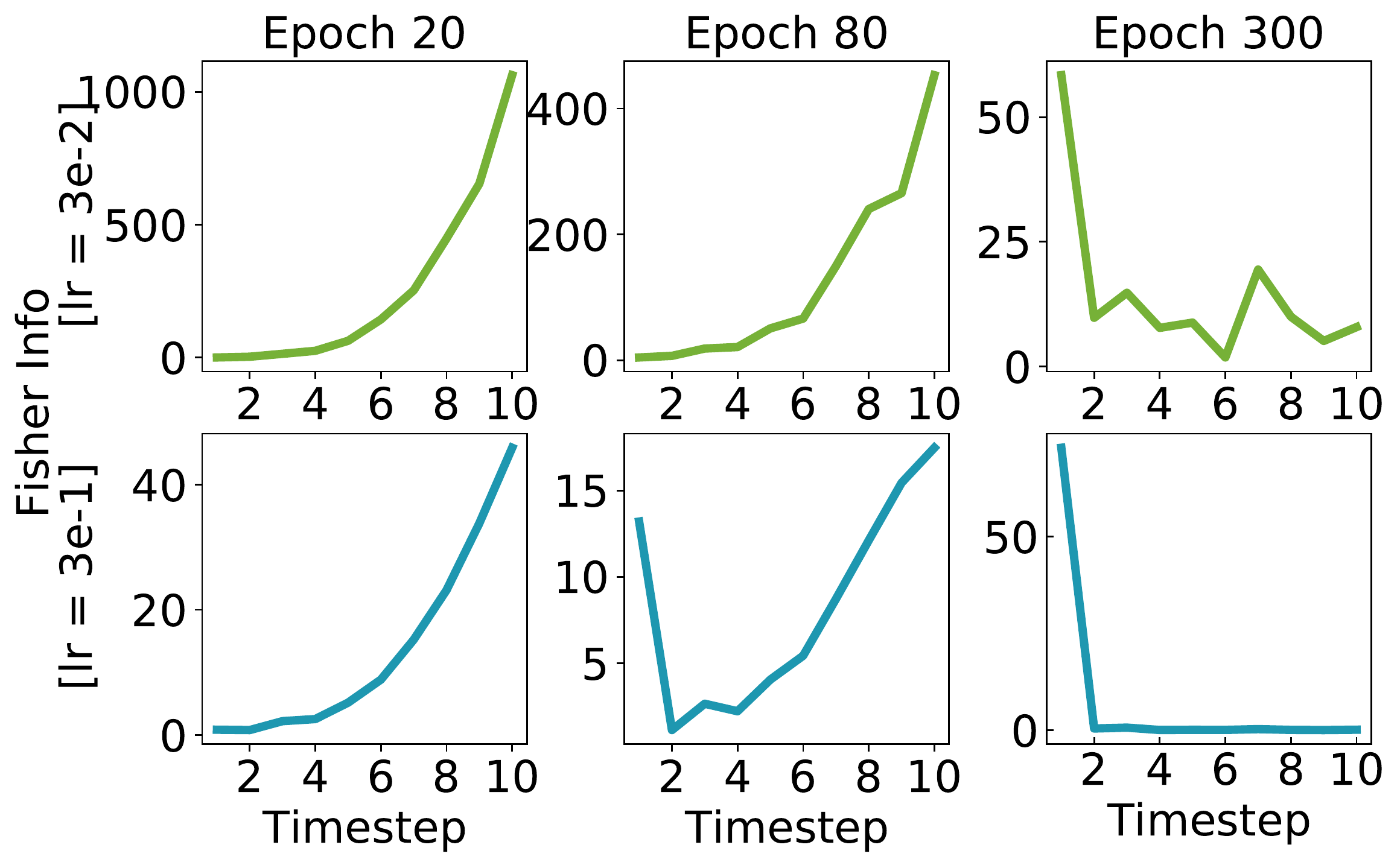} \\
{\hspace{1mm} (c) Weight decay} & {\hspace{1mm} (d) Learning rate} \\
 \includegraphics[width=0.30\linewidth]{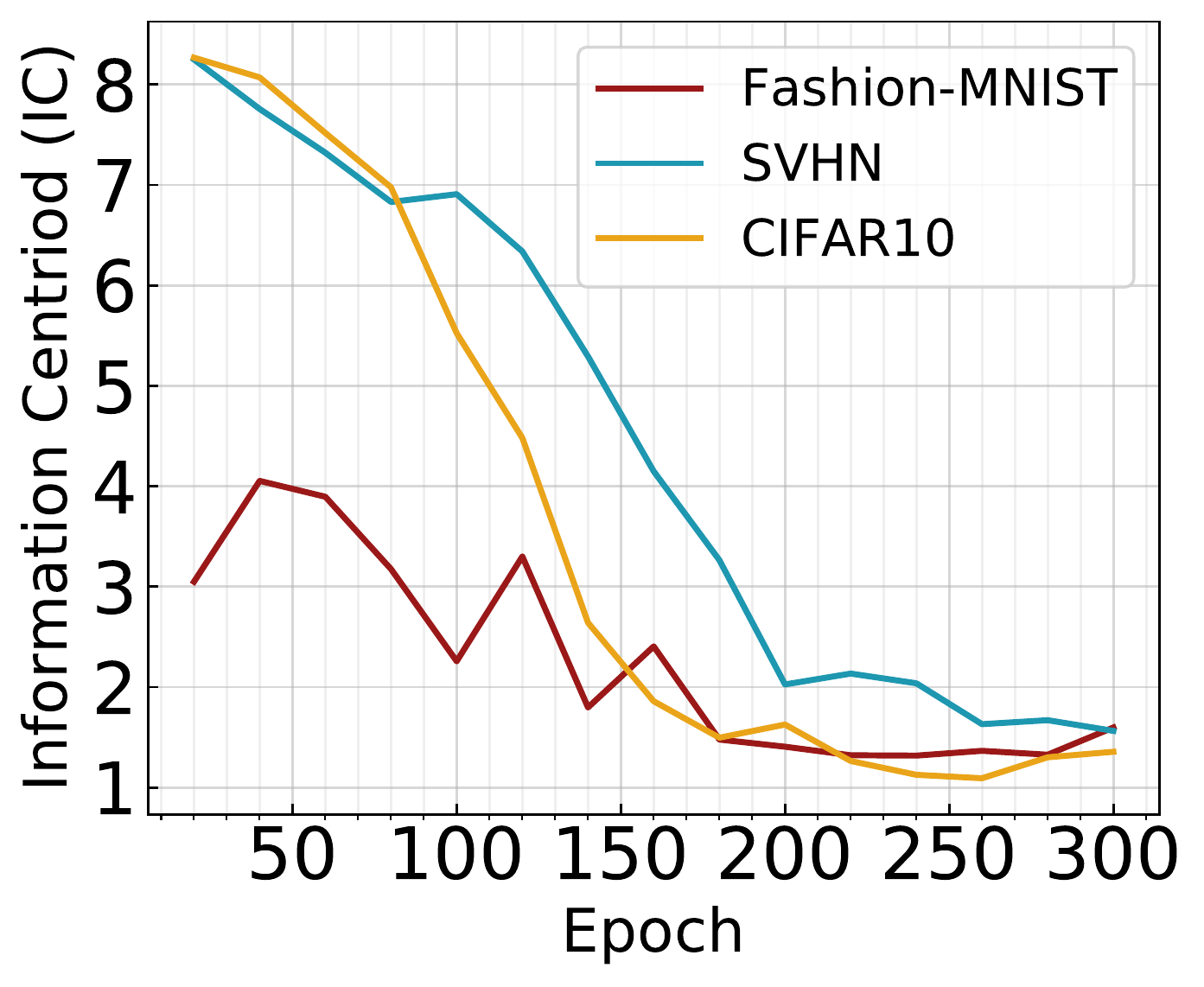} &
\includegraphics[width=0.30\linewidth]{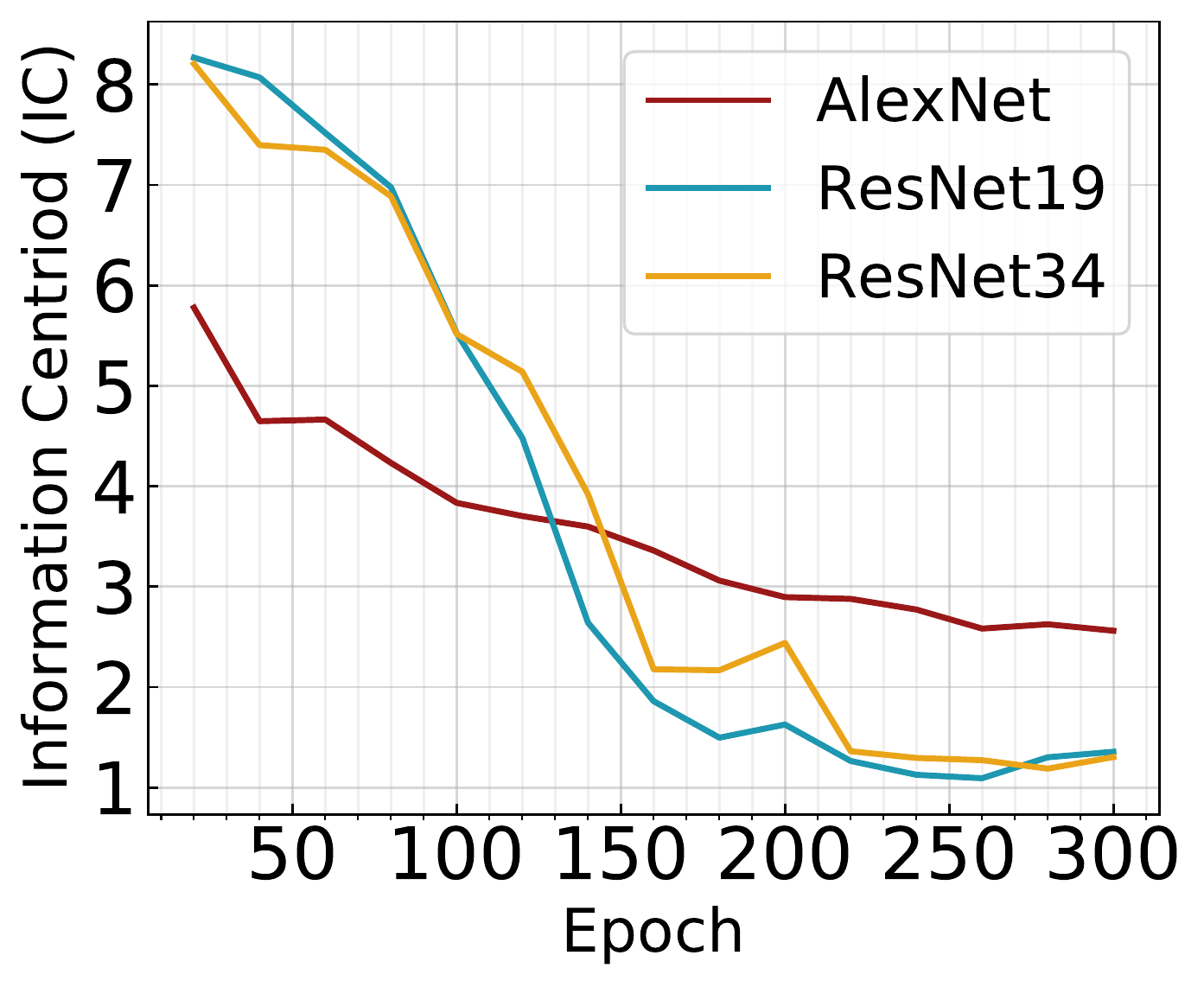} \hspace{-2mm}
\includegraphics[width=0.38\linewidth]{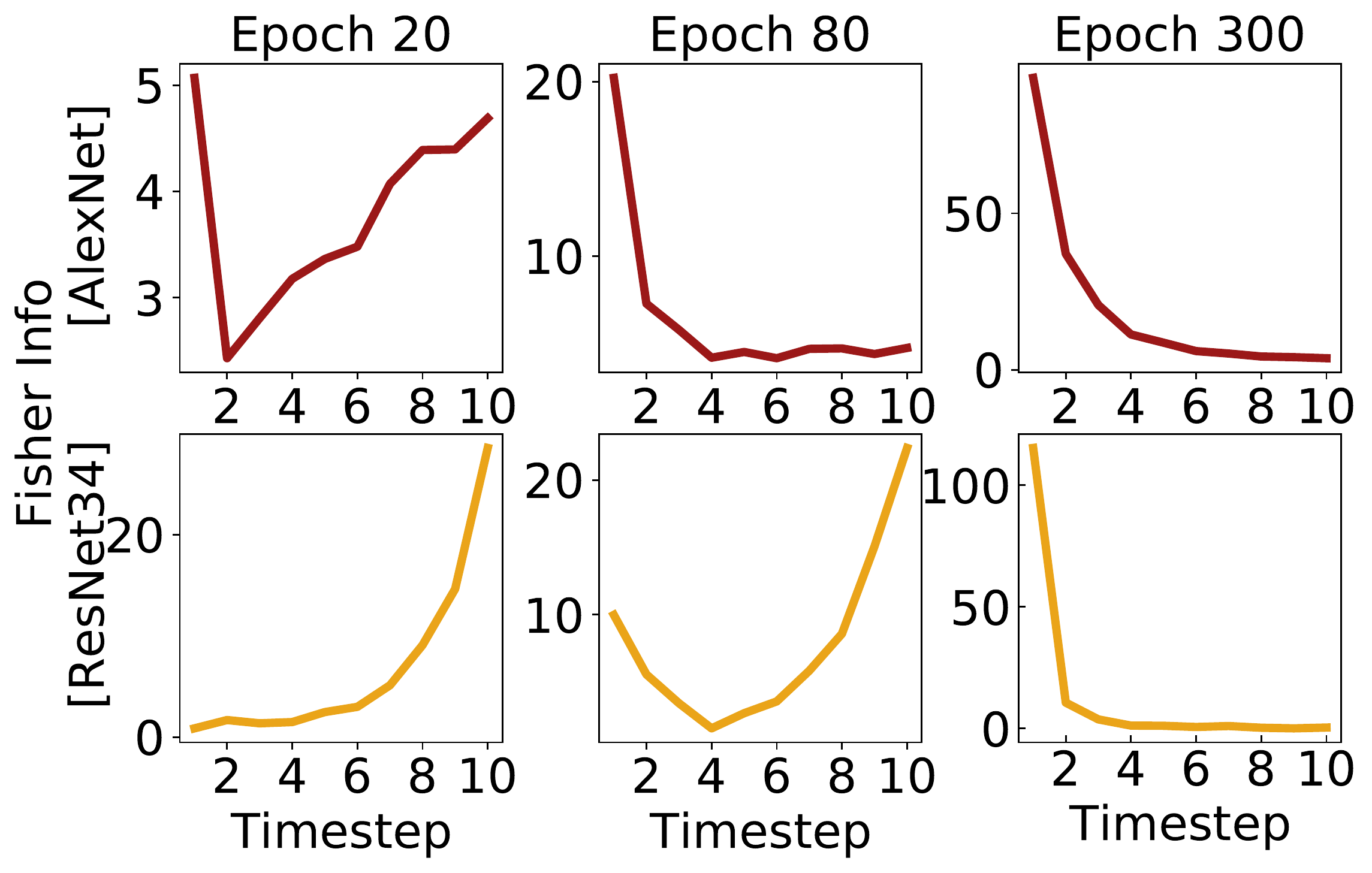}
\\
 \vspace{-2.5mm}
{\hspace{1mm} (e) Dataset} & {(f) Architecture}\\
\end{tabular}
\end{center}
\caption{ 
Information Centriod (IC) change with different factors. We observe that most cases show  TIC at the later epochs, where the early timesteps show higher Fisher information. For (b), (d), (f), in the right panels, we present detailed temporal Fisher information change at epoch 20, 80, and 300.
}
 \vspace{-4mm}
\label{fig:method:fisher_ablation}
\end{figure}

\textbf{Is temporal information concentration trend always shown in  SNNs?} 
 Having observed that TIC emerges in SNNs, we next study whether such phenomenon can be varied by various configurations such as different timesteps, time constant, learning rate, dataset and different architectures. We visualize the change of Information Centroid (IC) across epochs in Fig. \ref{fig:method:fisher_ablation}.
Additionally, for time constant, learning rate, architecture experiments (Fig. \ref{fig:method:fisher_ablation}(b), Fig. \ref{fig:method:fisher_ablation}(d), and Fig. \ref{fig:method:fisher_ablation}(f)), we also visualize timestep-wise Fisher information in order to provide more details in the following subsections.
For other experiments (Fig. \ref{fig:method:fisher_ablation}(a), Fig. \ref{fig:method:fisher_ablation}(c), and Fig. \ref{fig:method:fisher_ablation}(e)), we provide timestep-wise Fisher information in Appendix.
The default setting for all experiments is as follows: timestep 10, time constant 2, SGD optimizer with learning rate 3e-1, weight decay 5e-4, CIFAR10 dataset, and ResNet19 architecture.

\textit{Timestep:} We first train SNN on CIFAR10 with  timesteps 4, 6, 8, and 10.
We observe that all timestep configurations show a similar trend, \ie decreasing IC values as training goes on.
Note that longer timesteps have a higher initial IC value at the beginning of training.
One interesting observation is that a longer timestep starts late IC transition from high to low. For example, IC value of t=4 (green curve) starts to drop fast at the very early epochs, however, t=10 (red curve) shows IC transition near epoch 80.
This implies that a longer timestep contains more information, thus requires more training epochs to concentrate them in early timesteps.

\textit{Time constant:} As shown in Eq. \ref{eq:LIF}, time constant  $\tau$ controls the decaying intensity of the membrane potential of an LIF neuron.
A higher time constant means that the LIF neuron relies more on the previous information rather than the current input.
To evaluate the impact of time constant $\tau$ on TIC behavior, we train an SNN model with different time constants 0.75, 2, and 4.
The results show that the smaller time constants ($\tau=0.75$ and $\tau=2$) achieve low IC in the middle of training (see the yellow curves in the right panels of Fig. \ref{fig:method:fisher_ablation}(b)).
On the other hand, the higher  time constants ($\tau=4$) shows relatively higher IC compared to the others.
This is because, with high time constant, information is slowly propagated through layers, thus the information concentration begins in late timesteps (see the red curves in the right panels).

\textit{Weight decay and Learning rate:} We also analyze how learning rate affects TIC.
To this end, we use four different learning rate (lr) configurations: optimal lr (1e-3), lower lr (3e-2 and 8e-2), and higher lr (8e-1).
Compared to an optimal lr setting (blue curve), a lower lr (green and yellow curves) shows late IC transition from late timesteps to early timesteps.
Similarly, a larger lr (red curve) also does not show quick IC transition.
We also illustrate temporal Fisher information change for lr=3e-2 and lr=3e-1 in the right panels of Fig. \ref{fig:method:fisher_ablation}(d).
This results suggest that the early IC transition can be an indicator for choosing a proper learning rate in training process.
On the other hand, in Fig. \ref{fig:method:fisher_ablation}(c), we further conduct weight decay analysis, which shows weight decay does not make significant change on the temporal concentration behavior in SNN.

\textit{Dataset and Architecture:}
In Fig. \ref{fig:method:fisher_ablation}(e) and Fig. \ref{fig:method:fisher_ablation}(f), we show IC transition with different datasets and architectures.
For dataset ablation studies, we use CIFAR10~\cite{krizhevsky2009learning} and SVHN~\cite{netzer2011reading} that contain natural RGB images, as well as gray-scale Fashion-MNIST dataset~\cite{xiao2017fashion}.
All datasets show IC transition from late timesteps to early timesteps across training epochs.
Furthermore, we analyze on CNN architectures without skip connections (AlexNet \cite{krizhevsky2012imagenet}), with skip connections (ResNet19 \cite{he2016deep}), and deeper architecture (ResNet34 \cite{he2016deep}) with skip connections.  
AlexNet shows quick IC transition in the early epochs, however, it does not show very low IC value at the end of epochs.
This phenomenon also can be shown with temporal Fisher information visualization (right panels, Fig. \ref{fig:method:fisher_ablation}(f)) where AlexNet shows slow Fisher information decrease across time at epoch 300.
Different from AlexNet, ResNet34 shows quick information concentration in timesteps 1$\sim$3.
This suggests that AlexNet capacity is limited compared to ResNet34, therefore they require more timesteps to concentrate information.

Overall, we conclude that TIC is a common characteristic of SNNs, but IC transition speed can be  changed according to various factors such as optimization settings, architecture, and datasets.

\section{Analysis on TIC: Robustness Perspective}
\label{section:acc_robutness}
While the previous section shows temporal information concentration (TIC) is widely shared in SNNs, a key open question remains: \textit{what is the role of temporal concentration in SNN? does it bring higher accuracy or better robustness?} 
To find the answer, in this section, we force the SNN to have specific Fisher information trend across timesteps. In this way, we can investigate the characteristics and role of Fisher information in timesteps.

\begin{figure}[t]
\begin{center}
\def\arraystretch{0.5}
\begin{tabular}{@{}c@{\hskip 0.01\linewidth}c@{\hskip 0.01\linewidth}c@{}c@{}c@{}c}
\hspace{-4mm}
\includegraphics[width=0.331\linewidth]{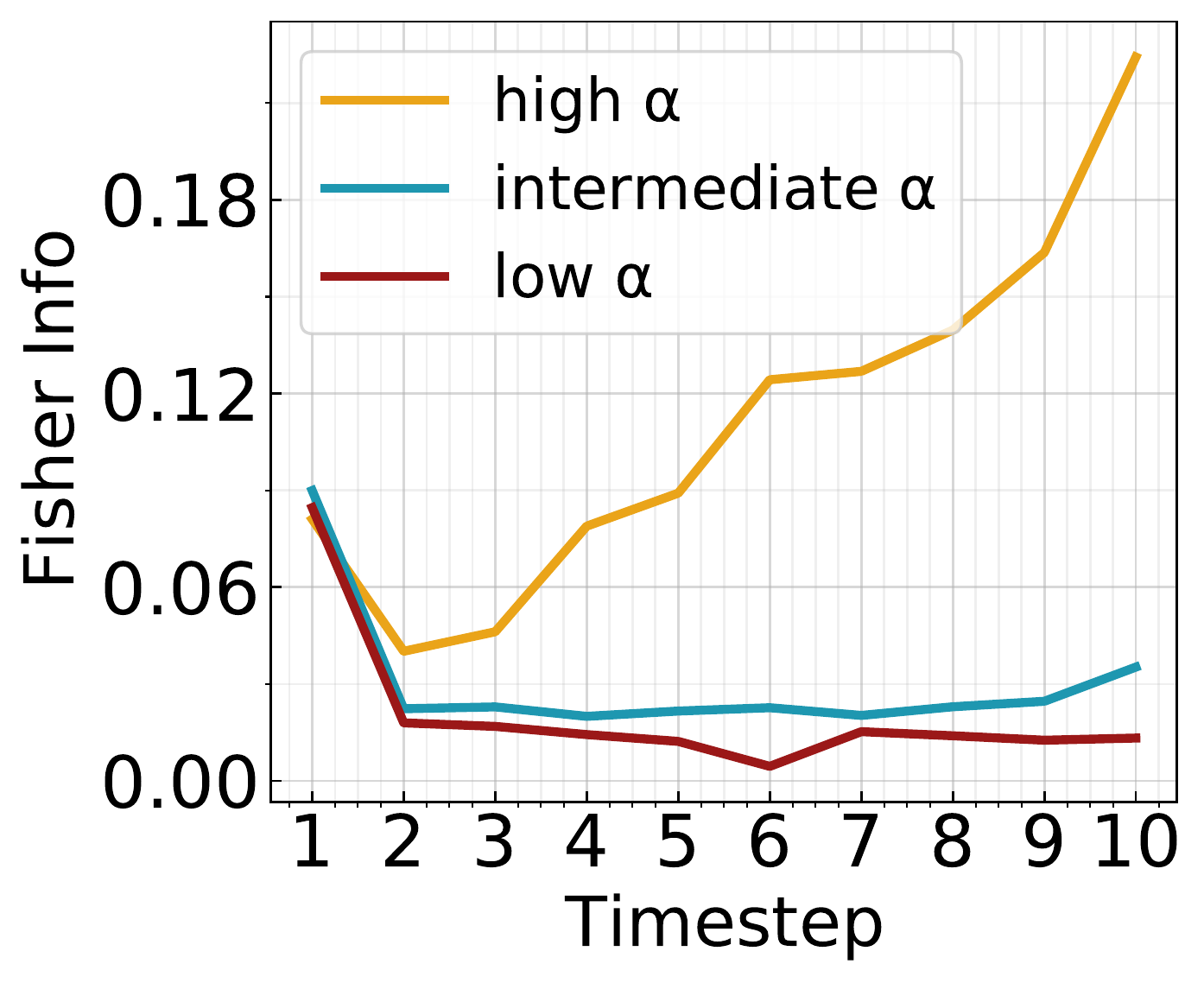} &
\includegraphics[width=0.333\linewidth]{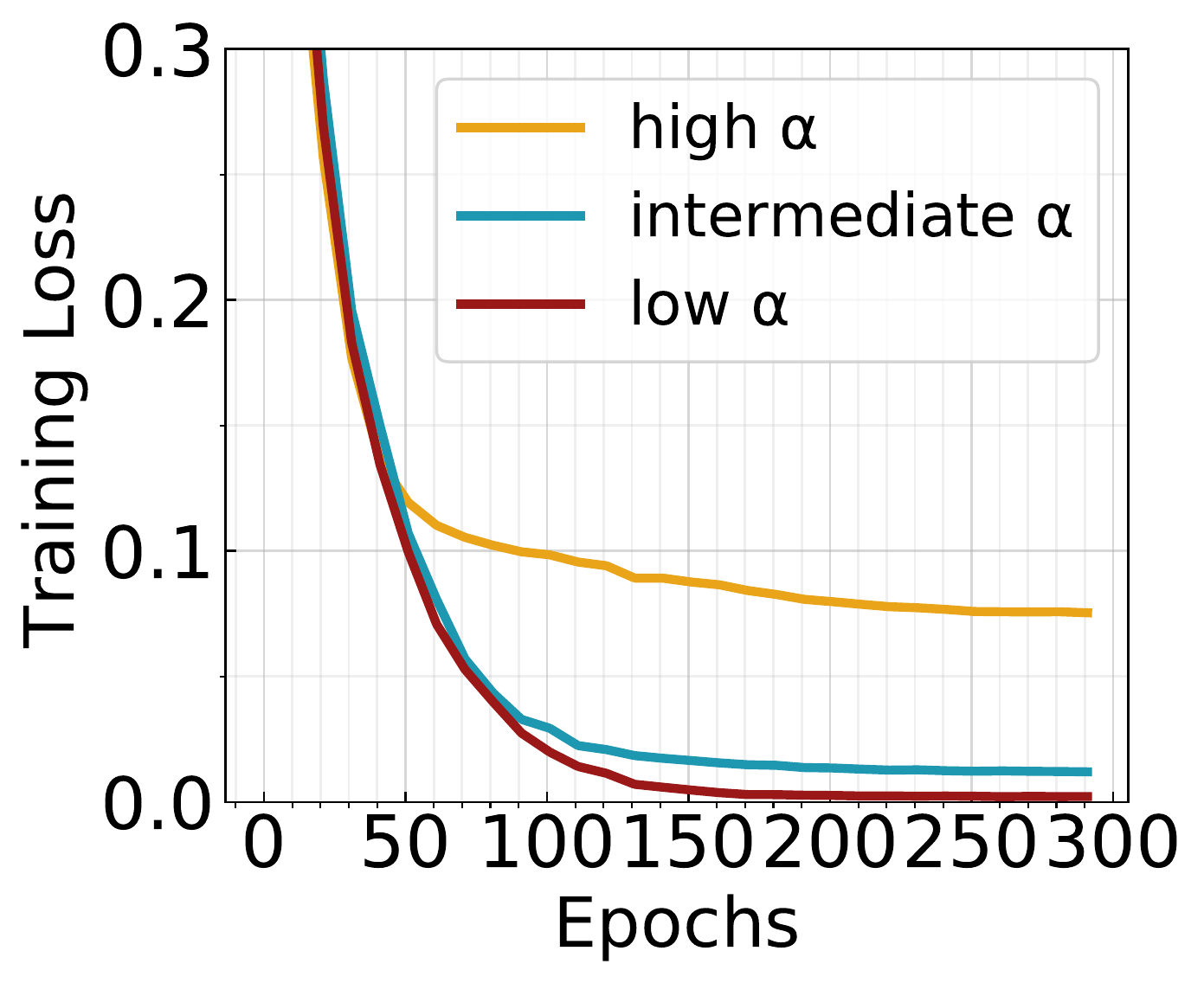} &
\includegraphics[width=0.333\linewidth]{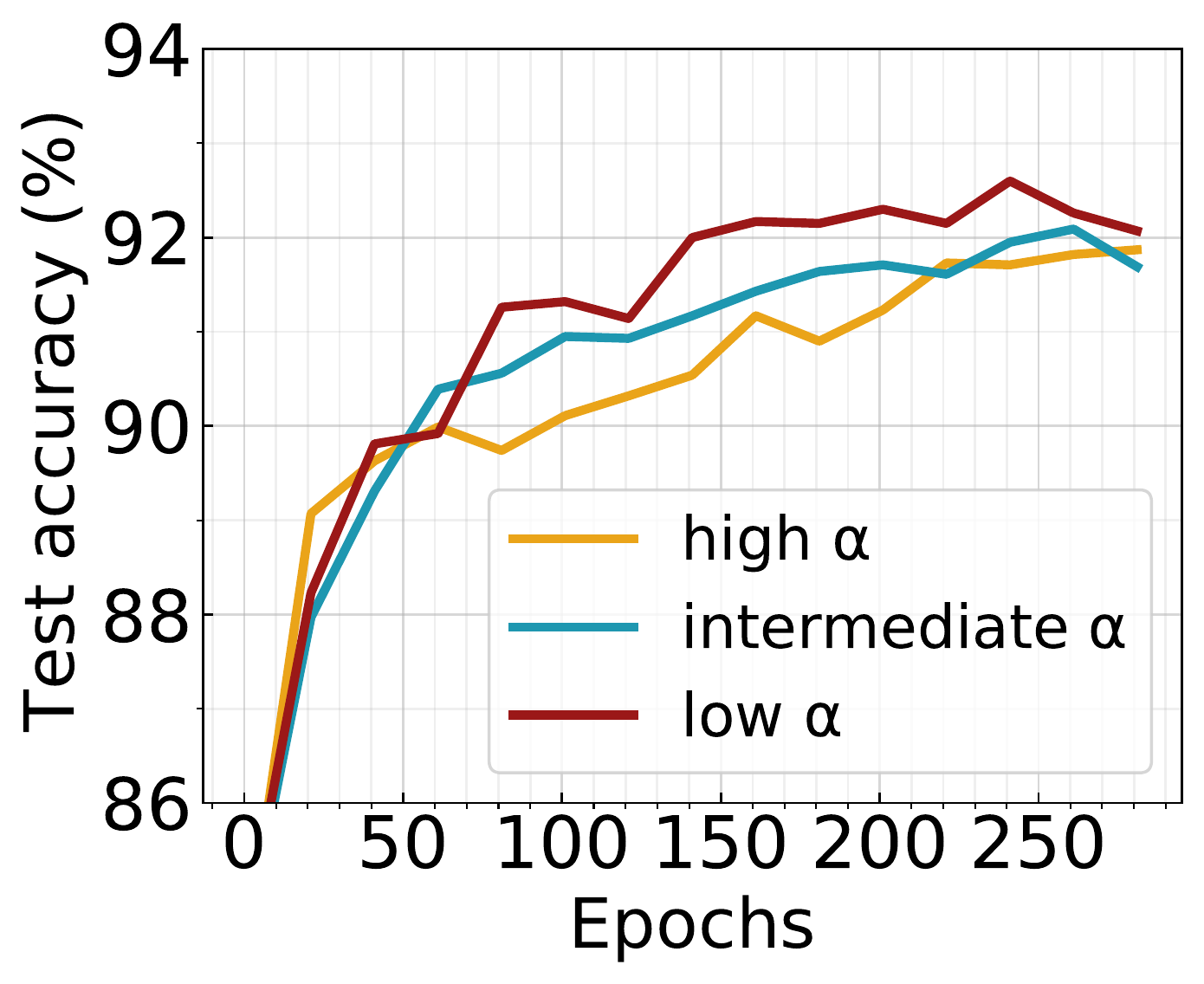}
\\
 \vspace{-2.5mm}
{\hspace{5mm}(a)} & {\hspace{5mm} (b)}  & {\hspace{5mm} (c)}\\
\end{tabular}
\end{center}
\caption{ (a) Fisher information across all layers after training (b) Training loss (c) Test accuracy of three $\alpha$ configurations. We train ResNet19 architecture on CIFAR10 dataset.
We observe that Fisher information successfully changes according to the magnitude of $\alpha$.
Also, (b) and (c) show our loss function provides stable convergence.
For loss visualization, we show the value of original loss $L_{t}({\theta})$ in Eq. \ref{eq:alphaL}.
We present the SVHN and CIFAR100 results in the Appendix.
}
\label{fig:method:loss_manipulation}
\end{figure}

To this end, we design a loss function to control the Fisher information trend in SNN.
Before designing the loss, we first define the approximated relation between a loss function and Fisher information, as shown in the following definition.
\begin{definition}
The log posterior  $log f_{\theta}(y|x_{i\le t})$ can be represented as a loss function $L_t({\theta})$, \eg cross-entropy loss, where the final layer's outputs are accumulated $t$ timesteps before they are converted to probabilities (\eg with a softmax layer). Thus, we can rewrite Eq. \ref{eq:final_information} as:
\begin{equation}
    I_{t}(\theta) = \mathbb{E}[\| \nabla L_{t}({\theta}) \|^2]. 
    \label{pr:eq:final_information}
\end{equation}
\label{definition1}
\end{definition}
Here, decreasing loss $L_t({\theta})$ will bring Fisher information $I_{t}(\theta)$ degradation because gradient $\nabla L_{t}({\theta})$ goes smaller as the loss converges to local minima.
On the other hand, if we disturb the model to converge a loss value, Fisher information cannot become smaller.
According to the aforementioned hypothesis, we control the  Fisher information value at each timestep, by manipulating the loss function during training.
Specifically, we force the loss function to have a value around $\alpha$:
\begin{equation}
    L_{t}({\theta}, \alpha) = | L_{t}({\theta}) - \alpha|.
    \label{eq:alphaL}
\end{equation}
The above equation represents that, if the loss function goes below $\alpha$, weights are updated with gradient ascent, otherwise a standard gradient descent is applied. Here, adding or subtracting $\alpha$ does not affect gradients for weight parameters. 
We apply Eq. \ref{eq:alphaL} across $T$ timesteps in order to make sure that Fisher information shows a similar trend for all timesteps, which can be formulated as: 
\begin{equation}
     L(\theta, \alpha) = \frac{1}{T} \sum_{t=1}^{T} L_{t}({\theta}, \alpha).
     \label{eq:loss_alphaexp}
\end{equation}
By changing $\alpha$ value, we can approximately control \textit{relative} magnitude of Fisher information. 
We find that our loss $L(\theta, \alpha)$ can successfully control Fisher information, \ie smaller $\alpha$ shows less Fisher information across timesteps,  as shown in Fig. \ref{fig:method:loss_manipulation}.
In our experiments, three types of SNNs trained with different  $\alpha$ values are investigated: $\alpha_{low}$,  $\alpha_{intermediate}$, and  $\alpha_{high}$ ($\alpha_{low} < \alpha_{intermediate} < \alpha_{high}$).
We select $\alpha_{cifar10}$=[1e-3, 1e-2, 7e-2], $\alpha_{svhn}$=[1e-4, 1e-2, 7e-2], $\alpha_{cifar100}$=[1e-4, 1e-3, 1e-2], for [$\alpha_{low}$,  $\alpha_{intermediate}$,   $\alpha_{high}$ ]. The $\alpha$ hyperparameter selection is based on dataset complexity where they have different sensitivity w.r.t to $\alpha$.

Compared to $\alpha_{low}$, using $\alpha_{intermediate}$ forces SNN to slightly increase the amount of Fisher information.
$\alpha_{high}$ further forces the model to have high Fisher Information, therefore the Fisher information increases as time goes on.
By comparing these configurations, we can reveal what is the advantage if Fisher information becomes smaller through time
\ie TIC. 
Note, the loss function (Eq. \ref{eq:loss_alphaexp}) applied for Fig. \ref{fig:method:loss_manipulation} and Table \ref{table:exp:acc_robust} is different from the other experiments where we only apply CrossEntropy loss on the accumulated spike activation in the last layer.
Our objective here is to manipulate Fisher information to explore their impact on the robustness of an SNN model.

\begin{table}[t]
\small
\centering
\resizebox{0.5\textwidth}{!}{%
\begin{tabular}{lc|c|cccc}
\toprule
Method  &  Dataset &  Clean Acc. (\%) & Gaussian Noise  & Blur & FGSM & PGD \\
\midrule
   low $\alpha$ & SVHN  & 96.03         & 93.48 (-2.55) &  95.56 (-0.47) & 91.69 (-4.34)	&  	90.87	(-5.16)  \\
    inter.  $\alpha$ & SVHN & 96.01&  93.24 (-2.77) &  95.56 (-0.56)   & 77.93	(-18.08)	&	49.84	(-46.17) \\
    high $\alpha$ & SVHN & 95.91         &  92.85 (-3.06)&  95.12 (-0.79)  & 55.39	(-40.52)	&	4.46	(-91.45)\\
    \midrule
     low $\alpha$ & CIFAR10 &  92.04  & 69.01 (-23.03)&  58.11 (-33.93) & 77.22 (-14.82) & 74.63 (-17.41)\\
    inter.  $\alpha$ & CIFAR10 &  91.89 &  68.09 (-23.80)&  56.88 (-35.01) & 69.29 (-22.60) & 58.65 (-33.24) \\
    high $\alpha$ & CIFAR10&  91.87  &  61.01 (-30.86)&  54.55 (-37.32) & 53.50 (-38.37) & 32.58 (-59.29) \\
    \midrule
     low $\alpha$ & CIFAR100 &  68.17  & 37.60 (-30.57) &  51.18	(-16.99) & 44.62 (-23.55) & 38.64 (-29.52)\\
    inter.  $\alpha$ & CIFAR100 &  68.47 &  36.98	(-31.49) &  51.02 (-17.45)  & 43.39	(-25.08) & 35.03	(-33.43) \\
    high $\alpha$ & CIFAR100&  67.95  &  35.56 (-32.39) &  49.09 (-18.86) & 38.36	(-29.59) & 30.33 (-37.62)\\
\bottomrule
\end{tabular}%
}
\caption{ Classification accuracy and robustness of SNNs trained with three different $\alpha$.  We train ResNet19 architecture on three public datasets including SVHN, CIFAR10, CIFAR100. For robustness experiments, we report both accuracy and relative accuracy drop w.r.t. clean accuracy. 
}
\label{table:exp:acc_robust}
  \vspace{-3mm}
\end{table}

\textbf{Classification Accuracy.} We first measure the accuracy of SNNs trained with three different  $\alpha$ (with the same number of epochs) on CIFAR10, CIFAR100~\cite{krizhevsky2009learning}, and SVHN~\cite{netzer2011reading}, and report the results in Table \ref{table:exp:acc_robust}. We use ResNet19 as a baseline architecture.
All $\alpha$ configurations achieve similar accuracy across all datasets, regardless of dataset complexity. 
This implies that TIC is not an essential factor for SNNs to obtain high accuracy.
In fact, in the TIC ablation experiments in  Fig. \ref{fig:method:fisher_ablation}, we found that IC transition speed changes based on different configurations, but, there was no conspicuous effect on accuracy. Table \ref{table:exp:acc_robust} results on the relation between TIC and accuracy further corroborate Fig. \ref{fig:method:fisher_ablation} (except for architecture and data experiments) results, where we achieve almost similar accuracy for all configurations.

\textbf{Robustness against  Noise / Adversarial sample.}
We analyze the relation between robustness and TIC.
Specifically, we first corrupt the input image using two types of noise: Gaussian noise and Blur.
For Gaussian noise experiments, we add Gaussian noise whose $l_2$ norm is set to 50\% of the norm of the given input.
For generating Blur effect to the image, we follow the method used in \cite{achille2018critical} - images are downsampled to smaller resolution (\eg $16\times16$) and then upsampled back to its original resolution (\eg $32\times32$) with bilinear interpolation, which is an efficient implementation for destroying small-scale details in images.
To further evaluate the robustness, we also evaluate the robustness of SNN models against two representative adversarial attacks. %
FGSM attack \cite{goodfellow2014explaining} is a single-step attack based on backward gradients where the noise is the sign of gradients scaled by $\epsilon$.
{Projected Gradient Descent (PGD)} attack \cite{madry2017towards} is an iterative adversarial attack characterized by three parameters- maximum perturbation $\epsilon$, perturbation step size $\alpha$ and the number of iterations $n$. 
In our experiments, we use $\epsilon = \frac{8}{255}$ for FGSM attack, and
[ $\epsilon = \frac{8}{255}$, $\alpha = \frac{4}{255}$, $n=10$] for PGD attack.

In Table \ref{table:exp:acc_robust}, we report the corrupted accuracy with respect to each noise.
Interestingly, for all noise configuration, SNNs trained with low $\alpha$ show less performance drop compared to SNNs trained with high $\alpha$.
This difference is especially huge for adversarial attacks.
This results can be explained by connecting the KL divergence to Fisher information matrix (FIM).
Given that a small perturbation $\delta$ is added to input $x$, it will make a difference in the probability $f_{\theta}(y|x_{t}+\delta)$.
Then, we can measure the difference in the output probabilities using KL divergence, where we can apply second-order Taylor approximation as
\vspace{-1mm}
\begin{equation}
    D_{KL}(f_{\theta}(y|x_{t})||f_{\theta}(y|x_{t}+\delta)) \approx \frac{1}{2} \delta^{T}M_{t}\delta.
\end{equation}
The above equation shows that the output perturbation from the given input noise can be represented by the function of FIM. If the eigen values  of the FIM have larger values, the output perturbation is likely to be severe.
Thus, the trace of FIM (\ie sum of eigen values) should be small in order to suppress the noise.
Therefore, the SNN model that shows temporal concentration behavior (smaller Fisher trace as time goes on) might have better robustness.

\begin{figure}[t]
\begin{center}
\def\arraystretch{0.5}
\begin{tabular}{@{}c@{\hskip 0.01\linewidth}c@{\hskip 0.01\linewidth}c@{}c@{}c@{}c}
\includegraphics[width=0.34\linewidth]{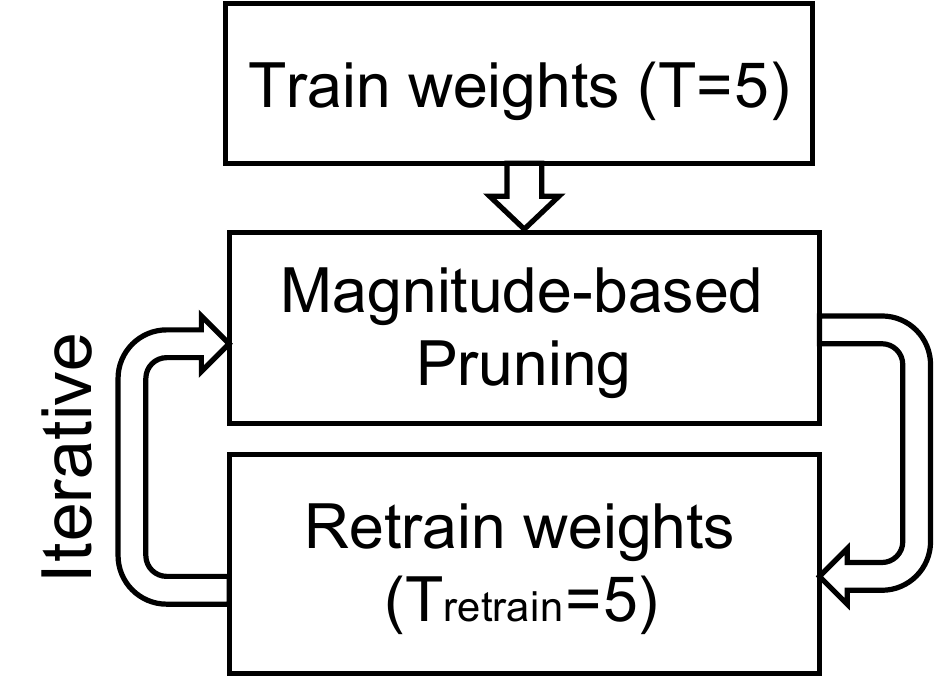} &
\includegraphics[width=0.65\linewidth]{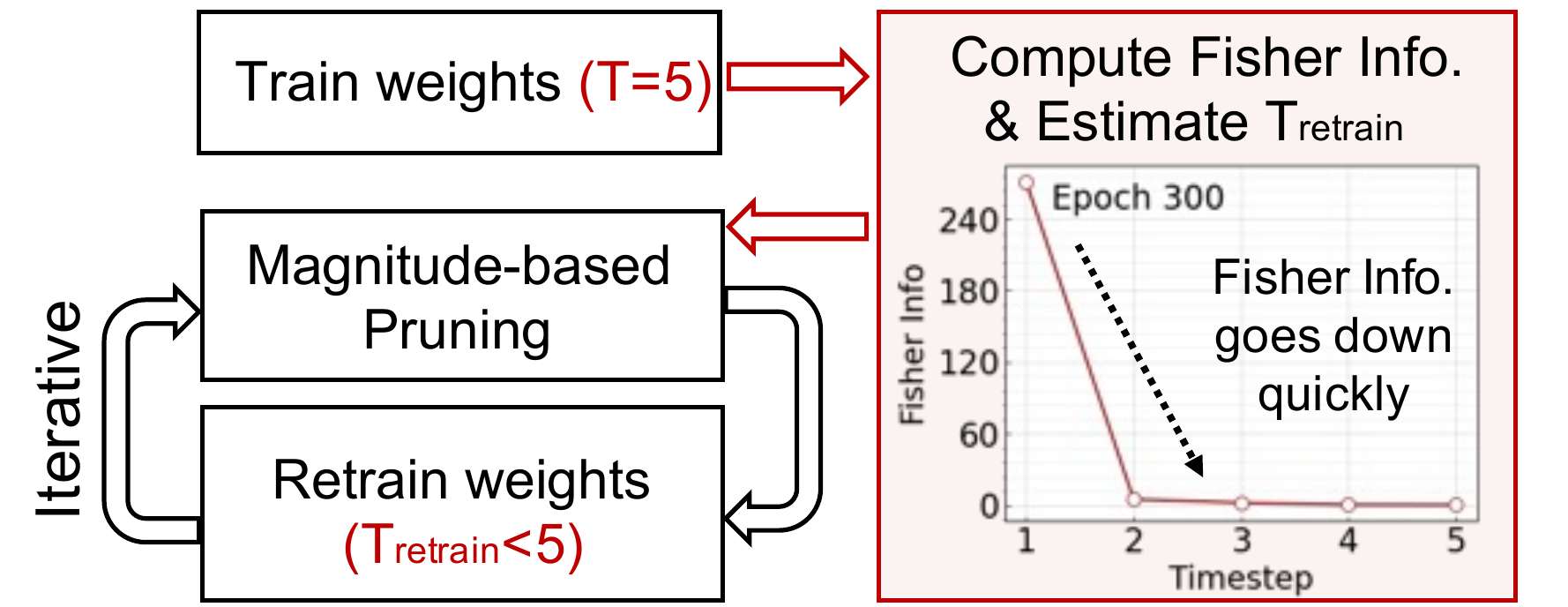} 
\\
{(a)} & { (b)} \\
 \vspace{-3.5mm}
\end{tabular}
\end{center}
\caption{ (a) Iterative pruning method \cite{han2015learning} for SNNs. Here, we assume the SNN are trained with timestep 5.  (b) We propose the concept of efficient pruning using TIC (colored with \textcolor{red}{red}). For a retraining-pruning cycle, the SNN model is trained with a less number of timesteps.
}
\label{fig:exp:pruning_methods}
\end{figure}

\begin{figure}[t]
\begin{center}
\def\arraystretch{0.5}
\begin{tabular}{@{}c@{\hskip 0.01\linewidth}c@{\hskip 0.01\linewidth}c@{}c@{}c@{}c}
\hspace{-4mm}
\includegraphics[width=0.45\linewidth]{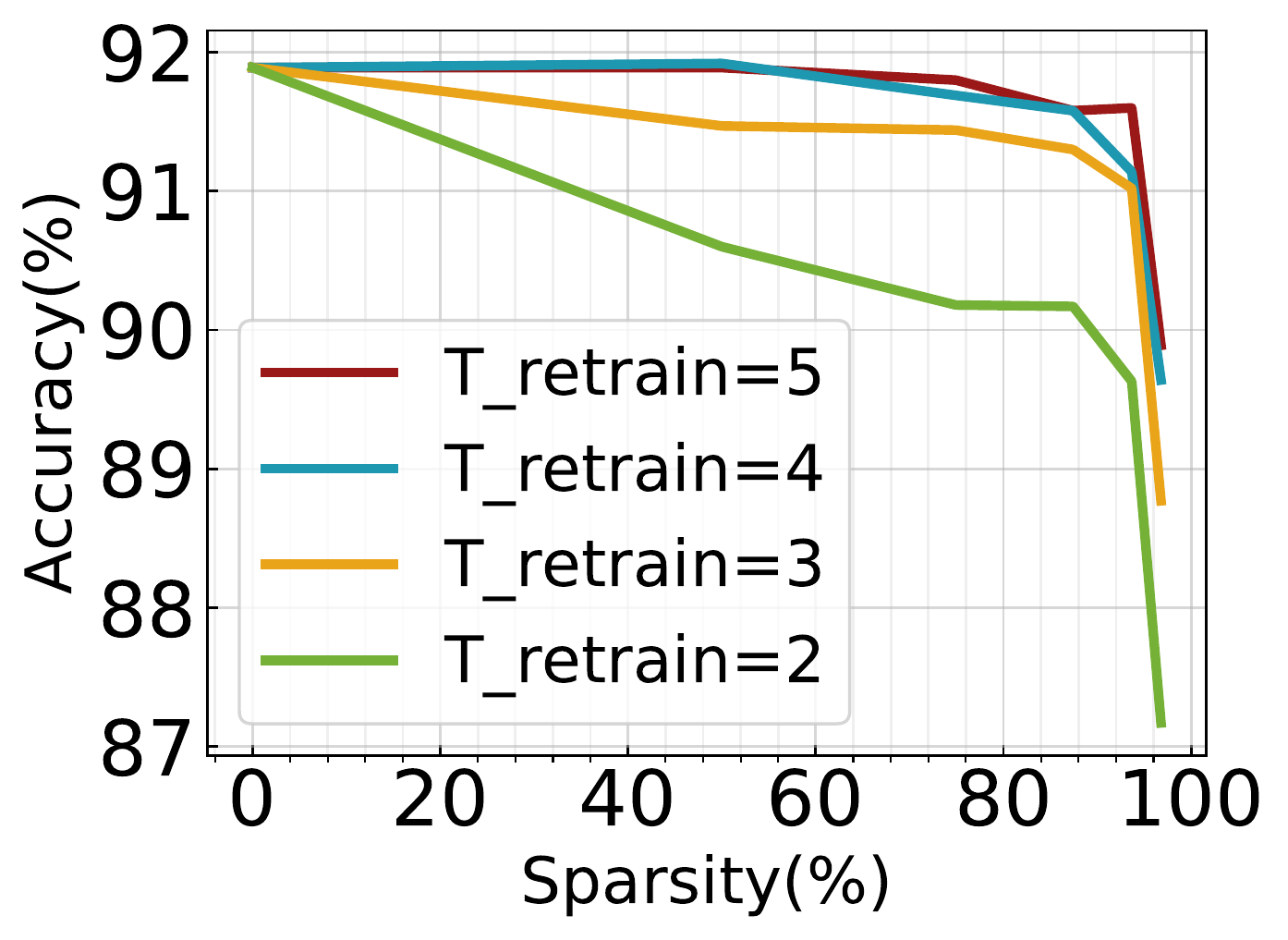} &
\includegraphics[width=0.45\linewidth]{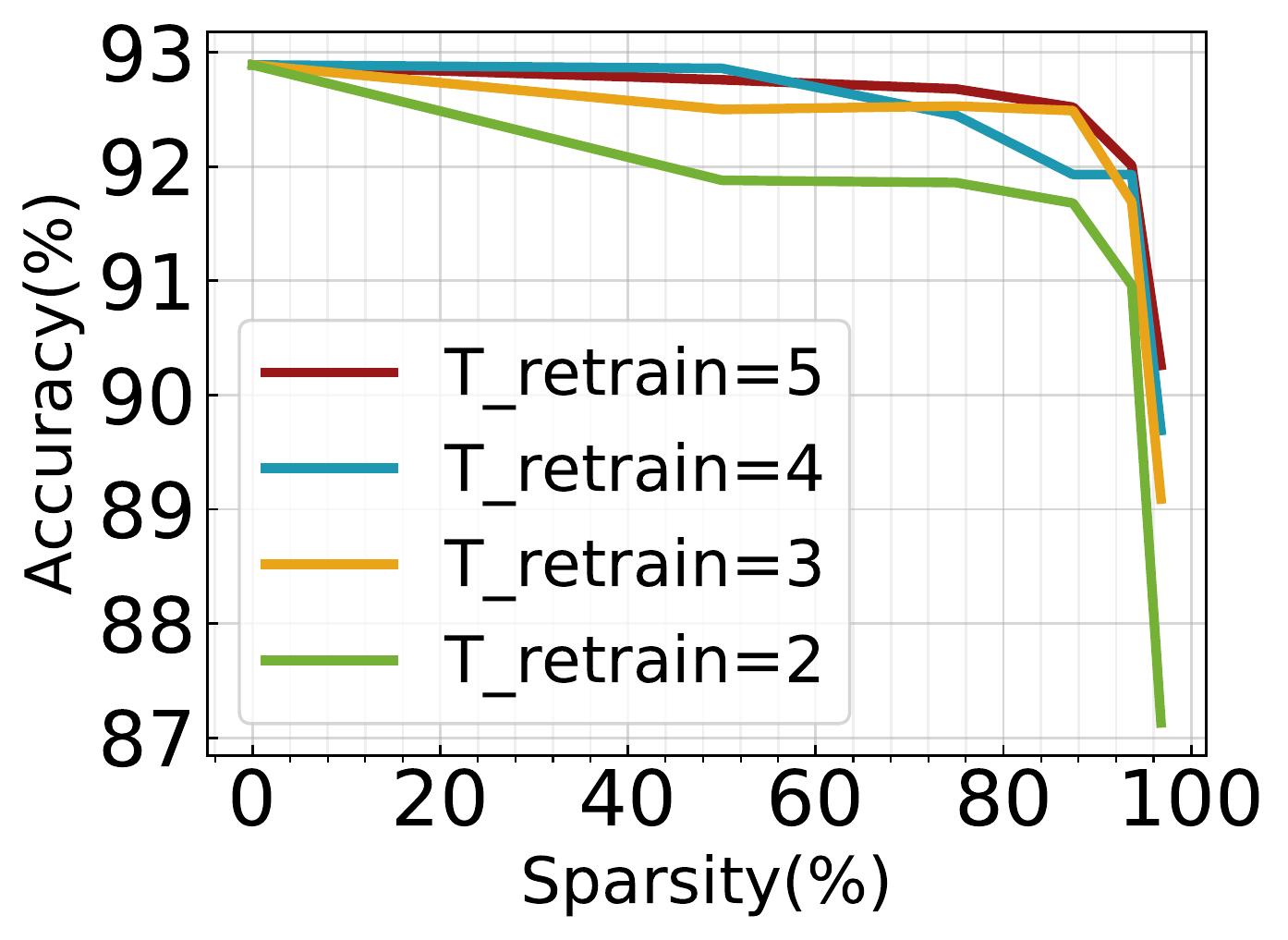} 
\\
 \vspace{-2.5mm}
{(a)} & { (b)} \\
\end{tabular}
\end{center}
\caption{ Accuracy vs. sparsity of (a) VGG16 (b) ResNet19 architectures on CIFAR10. In these experiments, we use timestep 5 for the first training stage, then we change $T_{retrain}$ in retraining-pruning cycles. 
}
  \vspace{-2mm}
\label{fig:exp:pruning_results}
\end{figure}

\section{Application: Efficient SNN Pruning using TIC}

We further explore the usage of TIC for applications.
Here, we apply a standard iterative magnitude-based pruning \cite{han2015learning} to SNNs and propose an efficient pruning process using TIC.
Basically, as shown in Fig. \ref{fig:exp:pruning_methods}(a), the original pruning method \cite{han2015learning} starts with training SNNs using timestep T. After training, $p\%$ of low-magnitude weight connections are pruned by thresholding, followed by retraining the remaining parameters. It was found that the iterative pruning strategy (\ie multiple pruning-retraining cycles) brings better pruning results. Following this, in our experiments, we prune 50\% of connections for 5 cycles.
As such iterative pruning causes non-trivial training time, we aim to reduce the computational cost using the TIC phenomenon in SNNs.

To bring efficiency, we focus on the observation that the information is concentrated in early timesteps after the first training stage.
In Fig. \ref{fig:exp:pruning_methods}(b), we illustrate the amount of Fisher information w.r.t timesteps, which shows near-zero values after T=3. 
This allows us to use less number of timesteps $T_{retrain}$ for pruning-retraining cycles since we will not lose information with a smaller number of timesteps.
In Fig. \ref{fig:exp:pruning_results}, we compute the accuracy across various sparsity levels with different $T_{retrain}$ values. 
For $T_{retrain}$ that shows near-zero Fisher information values (\ie $T_{retrain}$=[3, 4, 5]), we achieve similar pruning performance.
On the other hand, $T_{retrain}$=2 shows huge performance degradation.

Using small $T_{retrain}$ brings energy-efficiency.
Suppose we train the SNN model for $N$ epochs in the first training round, and $N_{retrain}$ epochs for $R$ rounds of the pruning-retraining process.
With the original approach \cite{han2015learning}, the computational cost is approximately proportional to $NT + N_{retrain}RT$, where $T$ is the number of timesteps.
If we apply $T_{retrain}$ for the pruning-retraining process based on TIC observation, the computational cost is approximately proportional to $NT + N_{retrain}RT_{retrain}$.
Overall, the compute efficiency can be obtained as: $\frac{N_{retrain}R(T-T_{train})}{NT + N_{retrain}RT} \times 100$.
In our experiment, we set $N=300$, $N_{retrain}=60$, $R=5$, $T=5$. In this case, if we apply $T_{retrain}=3$, we will obtain $20\%$ compute efficiency improvement.

\section{Discussion and Conclusion}
\label{seciton:discussion}
Spiking Neural Networks (SNNs) have recently gained great attention because of their interesting features similar to biological neurons  \cite{roy2019towards,diehl2015unsupervised,comsa2020temporal,mostafa2017supervised,sengupta2019going} and energy-efficiency. 
\textit{Although the previous works state that SNN may utilizes the temporal dynamics} \cite{wu2018spatio,fang2021incorporating,kim2020revisiting,masquelier2011timing}, \textit{yet, there is no prior work that conducts analysis on the information dynamics through the time dimension in SNN}.
In our work, we observe {temporal information concentration (TIC)} in SNNs, an information shift from latter timesteps to the early timesteps as training progresses.
This new observation enables us to understand the learning representations inside SNNs, providing several discussions on the connection to previous studies and future research directions for SNNs.

\textbf{Connection with bio-plausibility of SNNs.}
Interestingly, TIC reveals some connections between SNNs and biological features observed in human brain. 
Firstly, TIC also can be founded in the human visual cortex. Previous neuroscience research \cite{stigliani2017encoding, boynton1996linear} presented that the V1 activity responses to stimuli increase steeply at the early time and show little change afterward.
Also, layer-wise Fisher analysis (Fig. \ref{fig:method:layer_prop}) implies that  the SNN model focuses on high-level features (\eg semantic meaning of the given image) at the early timesteps, while low-level features (\eg texture) are important at the latter timesteps for prediction.
Such new observation in SNN is similar with primate vision where they focus on the semantic context of a given image at the early time, and then, look into finer details \cite{brady2009detecting,hollingworth2002accurate}.
Further, our observation raises several questions for neuroscience that could be studied using SNNs as a computational tool, such as,  how much of the early time visual stimuli is important in the learning period for humans? If there is noise at the early processing of visual stimuli, can humans recognize images correctly?

\textbf{Shrinking timesteps in SNN.}
Several works \cite{chowdhury2021one,chowdhury2021spatio} progressively reduce the number of timesteps as training goes on.
Although those works significantly reduce the latency of SNN while almost maintaining accuracy, there is a lack of explanation why timestep shrinking is possible in SNN.
Based on our observation, we can explain that SNNs can process most information in the early few timesteps, therefore achieving a high accuracy without the later timesteps.  
We hope our observation (\ie TIC, Fig. \ref{fig:method:fisher_train_base}) leads to interpretation of other temporal-related techniques in SNNs such as temporal batch norm \cite{zheng2020going,kim2020revisiting} and learnable leak factor \cite{fang2021incorporating}.

\textbf{Connection between model capacity and timesteps.}
In ANN literature, it is a well-known trend that a larger model  usually achieves better accuracy \cite{alzubaidi2021review,dosovitskiy2020image}.
Similarly, SNN works \cite{hu2018spiking,fang2021deep} also present that larger architectures achieve a better performance, showing that model capacity plays an important role in SNNs.
As our work focuses on understanding the relation between temporal dynamics and model learning, we conjecture that model capacity might affect the number of timesteps required for stable training.
Such hypothesis can be supported by observations in Fig. \ref{fig:method:fisher_ablation}(f), where high-capacity model (ResNet19) shows quick TIC compared to low-capacity model (AlexNet). 
Thus, the results imply high-capacity model can concentrate information within short timesteps.
To empirically validate this, as pilot experiments, we measured the accuracy of ResNet19 and AlexNet trained with various timesteps. Interestingly, we found that, ResNet19 shows performance saturation at timestep 4, whereas AlexNet saturates at timestep 8, which supports our statement that high-capacity model can be trained with short timesteps.
The results suggest that the trend of TIC can affect the minimum number of timesteps for performance saturation.
On the other hand, the final performance of model depends on the model capacity, where  AlexNet and ResNet achieve the accuracy of $\sim 80\%$ and $\sim 92\%$, respectively. 
We present experiment settings and results in Appendix.
Note, this observation is limited to only two models but suggests a new perspective on the model capacity and timesteps.

\textbf{Robustness of SNNs with respect to noise and adversarial attacks.}
Recent SNN works \cite{sharmin2020inherent,liang2021exploring} highlight that SNNs have better robustness against adversarial and natural noise compared to standard ANNs.
In our study, we further provide the analysis how temporal information dynamics affects robustness.
As we show that early timesteps are important for inference (thus, vulnerable), one can devise a more efficient attack/defense algorithm for SNN by adding noise at the early few timesteps.
We hope our study fosters future work understanding the robustness of SNNs.

\textbf{Impact of data type, loss function, coding scheme, and learning algorithm.}
In our experiments, we study the most general case of an SNN model (cross-entropy loss, direct coding, backpropagation for training) used in state-of-the-art works on image classification task.
We clarify that the trend TIC might be different with different SNN training configurations such as data type, loss function, coding scheme, and learning algorithm, which points out interesting future work to further understand SNNs.
Here we ask several research questions:
How information dynamics change with respect to sequential input such as DVS dataset \cite{calabrese2019dhp19,li2017cifar10}?
Will modifying cross-entropy loss in time axis help SNN training? 
Will different coding scheme (such as temporal coding \cite{mostafa2017supervised,comsa2020temporal}) show TIC trend?
Finally, can other SNN training algorithms based on time correlation, \eg Spike-timing-dependent plasticity (STDP) \cite{bi1998synaptic} preserve TIC trend?
We hope our work will become a starting point for thinking about a new direction in temporal representation for SNNs.

\section{{Acknowledgement}}  
This work was supported in part by C-BRIC, a JUMP center sponsored by DARPA and SRC, Google Research Scholar Award, the National Science Foundation (Grant\#1947826), TII (Abu Dhabi), the DARPA AI Exploration (AIE) program, and the DoE MMICC center SEA-CROGS (Award\#DE-SC0023198).

\bibliography{aaai23}

\begin{thebibliography}{60}
\providecommand{\natexlab}[1]{#1}

\bibitem[{Achille, Rovere, and Soatto(2018)}]{achille2018critical}
Achille, A.; Rovere, M.; and Soatto, S. 2018.
\newblock Critical learning periods in deep networks.
\newblock In \emph{International Conference on Learning Representations}.

\bibitem[{Akopyan et~al.(2015)Akopyan, Sawada, Cassidy, Alvarez-Icaza, Arthur,
  Merolla, Imam, Nakamura, Datta, Nam et~al.}]{akopyan2015truenorth}
Akopyan, F.; Sawada, J.; Cassidy, A.; Alvarez-Icaza, R.; Arthur, J.; Merolla,
  P.; Imam, N.; Nakamura, Y.; Datta, P.; Nam, G.-J.; et~al. 2015.
\newblock Truenorth: Design and tool flow of a 65 mw 1 million neuron
  programmable neurosynaptic chip.
\newblock \emph{IEEE transactions on computer-aided design of integrated
  circuits and systems}, 34(10): 1537--1557.

\bibitem[{Alzubaidi et~al.(2021)Alzubaidi, Zhang, Humaidi, Al-Dujaili, Duan,
  Al-Shamma, Santamar{\'\i}a, Fadhel, Al-Amidie, and
  Farhan}]{alzubaidi2021review}
Alzubaidi, L.; Zhang, J.; Humaidi, A.~J.; Al-Dujaili, A.; Duan, Y.; Al-Shamma,
  O.; Santamar{\'\i}a, J.; Fadhel, M.~A.; Al-Amidie, M.; and Farhan, L. 2021.
\newblock Review of deep learning: Concepts, CNN architectures, challenges,
  applications, future directions.
\newblock \emph{Journal of big Data}, 8(1): 1--74.

\bibitem[{Amari, Park, and Fukumizu(2000)}]{amari2000adaptive}
Amari, S.-i.; Park, H.; and Fukumizu, K. 2000.
\newblock Adaptive method of realizing natural gradient learning for multilayer
  perceptrons.
\newblock \emph{Neural computation}, 12(6): 1399--1409.

\bibitem[{Bi and Poo(1998)}]{bi1998synaptic}
Bi, G.-q.; and Poo, M.-m. 1998.
\newblock Synaptic modifications in cultured hippocampal neurons: dependence on
  spike timing, synaptic strength, and postsynaptic cell type.
\newblock \emph{Journal of neuroscience}, 18(24): 10464--10472.

\bibitem[{Boynton et~al.(1996)Boynton, Engel, Glover, and
  Heeger}]{boynton1996linear}
Boynton, G.~M.; Engel, S.~A.; Glover, G.~H.; and Heeger, D.~J. 1996.
\newblock Linear systems analysis of functional magnetic resonance imaging in
  human V1.
\newblock \emph{Journal of Neuroscience}, 16(13): 4207--4221.

\bibitem[{Brady et~al.(2009)Brady, Konkle, Oliva, and
  Alvarez}]{brady2009detecting}
Brady, T.~F.; Konkle, T.; Oliva, A.; and Alvarez, G.~A. 2009.
\newblock Detecting changes in real-world objects: The relationship between
  visual long-term memory and change blindness.
\newblock \emph{Communicative \& integrative biology}, 2(1): 1--3.

\bibitem[{Calabrese et~al.(2019)Calabrese, Taverni, Awai~Easthope, Skriabine,
  Corradi, Longinotti, Eng, and Delbruck}]{calabrese2019dhp19}
Calabrese, E.; Taverni, G.; Awai~Easthope, C.; Skriabine, S.; Corradi, F.;
  Longinotti, L.; Eng, K.; and Delbruck, T. 2019.
\newblock DHP19: Dynamic vision sensor 3D human pose dataset.
\newblock In \emph{Proceedings of the IEEE Conference on Computer Vision and
  Pattern Recognition Workshops}, 0--0.

\bibitem[{Chowdhury, Garg, and Roy(2021)}]{chowdhury2021spatio}
Chowdhury, S.~S.; Garg, I.; and Roy, K. 2021.
\newblock Spatio-Temporal Pruning and Quantization for Low-latency Spiking
  Neural Networks.
\newblock In \emph{2021 International Joint Conference on Neural Networks
  (IJCNN)}, 1--9. IEEE.

\bibitem[{Chowdhury, Rathi, and Roy(2021)}]{chowdhury2021one}
Chowdhury, S.~S.; Rathi, N.; and Roy, K. 2021.
\newblock One timestep is all you need: Training spiking neural networks with
  ultra low latency.
\newblock \emph{arXiv preprint arXiv:2110.05929}.

\bibitem[{Christensen et~al.(2022)Christensen, Dittmann, Linares-Barranco,
  Sebastian, Le~Gallo, Redaelli, Slesazeck, Mikolajick, Spiga, Menzel
  et~al.}]{christensen20222022}
Christensen, D.~V.; Dittmann, R.; Linares-Barranco, B.; Sebastian, A.;
  Le~Gallo, M.; Redaelli, A.; Slesazeck, S.; Mikolajick, T.; Spiga, S.; Menzel,
  S.; et~al. 2022.
\newblock 2022 roadmap on neuromorphic computing and engineering.
\newblock \emph{Neuromorphic Computing and Engineering}.

\bibitem[{Comsa et~al.(2020)Comsa, Fischbacher, Potempa, Gesmundo, Versari, and
  Alakuijala}]{comsa2020temporal}
Comsa, I.~M.; Fischbacher, T.; Potempa, K.; Gesmundo, A.; Versari, L.; and
  Alakuijala, J. 2020.
\newblock Temporal coding in spiking neural networks with alpha synaptic
  function.
\newblock In \emph{ICASSP 2020-2020 IEEE International Conference on Acoustics,
  Speech and Signal Processing (ICASSP)}, 8529--8533. IEEE.

\bibitem[{Davies et~al.(2018)Davies, Srinivasa, Lin, Chinya, Cao, Choday,
  Dimou, Joshi, Imam, Jain et~al.}]{davies2018loihi}
Davies, M.; Srinivasa, N.; Lin, T.-H.; Chinya, G.; Cao, Y.; Choday, S.~H.;
  Dimou, G.; Joshi, P.; Imam, N.; Jain, S.; et~al. 2018.
\newblock Loihi: A neuromorphic manycore processor with on-chip learning.
\newblock \emph{IEEE Micro}, 38(1): 82--99.

\bibitem[{Deng et~al.(2022)Deng, Li, Zhang, and Gu}]{deng2022temporal}
Deng, S.; Li, Y.; Zhang, S.; and Gu, S. 2022.
\newblock Temporal Efficient Training of Spiking Neural Network via Gradient
  Re-weighting.
\newblock \emph{arXiv preprint arXiv:2202.11946}.

\bibitem[{Diehl and Cook(2015)}]{diehl2015unsupervised}
Diehl, P.~U.; and Cook, M. 2015.
\newblock Unsupervised learning of digit recognition using
  spike-timing-dependent plasticity.
\newblock \emph{Frontiers in computational neuroscience}, 9: 99.

\bibitem[{Dosovitskiy et~al.(2020)Dosovitskiy, Beyer, Kolesnikov, Weissenborn,
  Zhai, Unterthiner, Dehghani, Minderer, Heigold, Gelly
  et~al.}]{dosovitskiy2020image}
Dosovitskiy, A.; Beyer, L.; Kolesnikov, A.; Weissenborn, D.; Zhai, X.;
  Unterthiner, T.; Dehghani, M.; Minderer, M.; Heigold, G.; Gelly, S.; et~al.
  2020.
\newblock An image is worth 16x16 words: Transformers for image recognition at
  scale.
\newblock \emph{arXiv preprint arXiv:2010.11929}.

\bibitem[{Fang et~al.(2021{\natexlab{a}})Fang, Yu, Chen, Huang, Masquelier, and
  Tian}]{fang2021deep}
Fang, W.; Yu, Z.; Chen, Y.; Huang, T.; Masquelier, T.; and Tian, Y.
  2021{\natexlab{a}}.
\newblock Deep Residual Learning in Spiking Neural Networks.
\newblock \emph{arXiv preprint arXiv:2102.04159}.

\bibitem[{Fang et~al.(2021{\natexlab{b}})Fang, Yu, Chen, Masquelier, Huang, and
  Tian}]{fang2021incorporating}
Fang, W.; Yu, Z.; Chen, Y.; Masquelier, T.; Huang, T.; and Tian, Y.
  2021{\natexlab{b}}.
\newblock Incorporating learnable membrane time constant to enhance learning of
  spiking neural networks.
\newblock In \emph{Proceedings of the IEEE/CVF International Conference on
  Computer Vision}, 2661--2671.

\bibitem[{Fisher(1925)}]{fisher1925theory}
Fisher, R.~A. 1925.
\newblock Theory of statistical estimation.
\newblock In \emph{Mathematical proceedings of the Cambridge philosophical
  society}, volume~22, 700--725. Cambridge University Press.

\bibitem[{Furber et~al.(2014)Furber, Galluppi, Temple, and
  Plana}]{furber2014spinnaker}
Furber, S.~B.; Galluppi, F.; Temple, S.; and Plana, L.~A. 2014.
\newblock The spinnaker project.
\newblock \emph{Proceedings of the IEEE}, 102(5): 652--665.

\bibitem[{Goodfellow et~al.(2014)}]{goodfellow2014explaining}
Goodfellow, I.~J.; et~al. 2014.
\newblock Explaining and harnessing adversarial examples.
\newblock \emph{arXiv preprint arXiv:1412.6572}.

\bibitem[{Han et~al.(2015)Han, Pool, Tran, and Dally}]{han2015learning}
Han, S.; Pool, J.; Tran, J.; and Dally, W. 2015.
\newblock Learning both weights and connections for efficient neural network.
\newblock \emph{Advances in neural information processing systems}, 28.

\bibitem[{He et~al.(2016)He, Zhang, Ren, and Sun}]{he2016deep}
He, K.; Zhang, X.; Ren, S.; and Sun, J. 2016.
\newblock Deep residual learning for image recognition.
\newblock 770--778.

\bibitem[{Hollingworth and Henderson(2002)}]{hollingworth2002accurate}
Hollingworth, A.; and Henderson, J.~M. 2002.
\newblock Accurate visual memory for previously attended objects in natural
  scenes.
\newblock \emph{Journal of Experimental Psychology: Human Perception and
  Performance}, 28(1): 113.

\bibitem[{Hu, Tang, and Pan(2018)}]{hu2018spiking}
Hu, Y.; Tang, H.; and Pan, G. 2018.
\newblock Spiking Deep Residual Networks.
\newblock \emph{IEEE Transactions on Neural Networks and Learning Systems}.

\bibitem[{Karakida, Akaho, and Amari(2019)}]{karakida2019universal}
Karakida, R.; Akaho, S.; and Amari, S.-i. 2019.
\newblock Universal statistics of fisher information in deep neural networks:
  Mean field approach.
\newblock In \emph{The 22nd International Conference on Artificial Intelligence
  and Statistics}, 1032--1041. PMLR.

\bibitem[{Keskar et~al.(2016)Keskar, Mudigere, Nocedal, Smelyanskiy, and
  Tang}]{keskar2016large}
Keskar, N.~S.; Mudigere, D.; Nocedal, J.; Smelyanskiy, M.; and Tang, P. T.~P.
  2016.
\newblock On large-batch training for deep learning: Generalization gap and
  sharp minima.
\newblock \emph{arXiv preprint arXiv:1609.04836}.

\bibitem[{Kim et~al.(2022{\natexlab{a}})Kim, Li, Park, Venkatesha, and
  Panda}]{kim2022neural}
Kim, Y.; Li, Y.; Park, H.; Venkatesha, Y.; and Panda, P. 2022{\natexlab{a}}.
\newblock Neural Architecture Search for Spiking Neural Networks.
\newblock \emph{arXiv preprint arXiv:2201.10355}.

\bibitem[{Kim et~al.(2022{\natexlab{b}})Kim, Li, Park, Venkatesha, Yin, and
  Panda}]{kim2022exploring}
Kim, Y.; Li, Y.; Park, H.; Venkatesha, Y.; Yin, R.; and Panda, P.
  2022{\natexlab{b}}.
\newblock Exploring Lottery Ticket Hypothesis in Spiking Neural Networks.
\newblock In \emph{European Conference on Computer Vision}, 102--120. Springer.

\bibitem[{Kim and Panda(2020)}]{kim2020revisiting}
Kim, Y.; and Panda, P. 2020.
\newblock Revisiting batch normalization for training low-latency deep spiking
  neural networks from scratch.
\newblock \emph{Frontiers in neuroscience}, 1638.

\bibitem[{Kim and Panda(2021)}]{kim2021visual}
Kim, Y.; and Panda, P. 2021.
\newblock Visual explanations from spiking neural networks using inter-spike
  intervals.
\newblock \emph{Scientific reports}, 11(1): 1--14.

\bibitem[{Kirkpatrick et~al.(2017)Kirkpatrick, Pascanu, Rabinowitz, Veness,
  Desjardins, Rusu, Milan, Quan, Ramalho, Grabska-Barwinska
  et~al.}]{kirkpatrick2017overcoming}
Kirkpatrick, J.; Pascanu, R.; Rabinowitz, N.; Veness, J.; Desjardins, G.; Rusu,
  A.~A.; Milan, K.; Quan, J.; Ramalho, T.; Grabska-Barwinska, A.; et~al. 2017.
\newblock Overcoming catastrophic forgetting in neural networks.
\newblock \emph{Proceedings of the national academy of sciences}, 114(13):
  3521--3526.

\bibitem[{Krizhevsky, Hinton et~al.(2009)}]{krizhevsky2009learning}
Krizhevsky, A.; Hinton, G.; et~al. 2009.
\newblock Learning multiple layers of features from tiny images.

\bibitem[{Krizhevsky, Sutskever, and Hinton(2012)}]{krizhevsky2012imagenet}
Krizhevsky, A.; Sutskever, I.; and Hinton, G.~E. 2012.
\newblock Imagenet classification with deep convolutional neural networks.
\newblock \emph{Advances in neural information processing systems}, 25.

\bibitem[{Lee et~al.(2020)Lee, Sarwar, Panda, Srinivasan, and
  Roy}]{lee2020enabling}
Lee, C.; Sarwar, S.~S.; Panda, P.; Srinivasan, G.; and Roy, K. 2020.
\newblock Enabling spike-based backpropagation for training deep neural network
  architectures.
\newblock \emph{Frontiers in Neuroscience}, 14.

\bibitem[{Li et~al.(2017)Li, Liu, Ji, Li, and Shi}]{li2017cifar10}
Li, H.; Liu, H.; Ji, X.; Li, G.; and Shi, L. 2017.
\newblock Cifar10-dvs: an event-stream dataset for object classification.
\newblock \emph{Frontiers in neuroscience}, 11: 309.

\bibitem[{Li et~al.(2021{\natexlab{a}})Li, Deng, Dong, Gong, and
  Gu}]{li2021free}
Li, Y.; Deng, S.; Dong, X.; Gong, R.; and Gu, S. 2021{\natexlab{a}}.
\newblock A Free Lunch From ANN: Towards Efficient, Accurate Spiking Neural
  Networks Calibration.
\newblock \emph{arXiv preprint arXiv:2106.06984}.

\bibitem[{Li et~al.(2021{\natexlab{b}})Li, Guo, Zhang, Deng, Hai, and
  Gu}]{li2021differentiable}
Li, Y.; Guo, Y.; Zhang, S.; Deng, S.; Hai, Y.; and Gu, S. 2021{\natexlab{b}}.
\newblock Differentiable Spike: Rethinking Gradient-Descent for Training
  Spiking Neural Networks.
\newblock \emph{Advances in Neural Information Processing Systems}, 34.

\bibitem[{Liang et~al.(2021)Liang, Hu, Deng, Wu, Li, Ding, Li, and
  Xie}]{liang2021exploring}
Liang, L.; Hu, X.; Deng, L.; Wu, Y.; Li, G.; Ding, Y.; Li, P.; and Xie, Y.
  2021.
\newblock Exploring adversarial attack in spiking neural networks with
  spike-compatible gradient.
\newblock \emph{IEEE transactions on neural networks and learning systems}.

\bibitem[{Liang et~al.(2019)Liang, Poggio, Rakhlin, and
  Stokes}]{liang2019fisher}
Liang, T.; Poggio, T.; Rakhlin, A.; and Stokes, J. 2019.
\newblock Fisher-rao metric, geometry, and complexity of neural networks.
\newblock In \emph{The 22nd international conference on artificial intelligence
  and statistics}, 888--896. PMLR.

\bibitem[{Lobo et~al.(2020)Lobo, Del~Ser, Bifet, and Kasabov}]{lobo2020spiking}
Lobo, J.~L.; Del~Ser, J.; Bifet, A.; and Kasabov, N. 2020.
\newblock Spiking neural networks and online learning: An overview and
  perspectives.
\newblock \emph{Neural Networks}, 121: 88--100.

\bibitem[{Ly et~al.(2017)Ly, Marsman, Verhagen, Grasman, and
  Wagenmakers}]{ly2017tutorial}
Ly, A.; Marsman, M.; Verhagen, J.; Grasman, R.~P.; and Wagenmakers, E.-J. 2017.
\newblock A tutorial on Fisher information.
\newblock \emph{Journal of Mathematical Psychology}, 80: 40--55.

\bibitem[{Madry et~al.(2017)}]{madry2017towards}
Madry, A.; et~al. 2017.
\newblock Towards deep learning models resistant to adversarial attacks.
\newblock \emph{arXiv preprint arXiv:1706.06083}.

\bibitem[{Martens and Grosse(2015)}]{martens2015optimizing}
Martens, J.; and Grosse, R. 2015.
\newblock Optimizing neural networks with kronecker-factored approximate
  curvature.
\newblock In \emph{International conference on machine learning}, 2408--2417.
  PMLR.

\bibitem[{Masquelier, Albantakis, and Deco(2011)}]{masquelier2011timing}
Masquelier, T.; Albantakis, L.; and Deco, G. 2011.
\newblock The timing of vision--how neural processing links to different
  temporal dynamics.
\newblock \emph{Frontiers in psychology}, 2: 151.

\bibitem[{Mostafa(2017)}]{mostafa2017supervised}
Mostafa, H. 2017.
\newblock Supervised learning based on temporal coding in spiking neural
  networks.
\newblock \emph{IEEE transactions on neural networks and learning systems},
  29(7): 3227--3235.

\bibitem[{Neftci, Mostafa, and Zenke(2019)}]{neftci2019surrogate}
Neftci, E.~O.; Mostafa, H.; and Zenke, F. 2019.
\newblock Surrogate gradient learning in spiking neural networks.
\newblock \emph{IEEE Signal Processing Magazine}, 36: 61--63.

\bibitem[{Netzer et~al.(2011)Netzer, Wang, Coates, Bissacco, Wu, and
  Ng}]{netzer2011reading}
Netzer, Y.; Wang, T.; Coates, A.; Bissacco, A.; Wu, B.; and Ng, A.~Y. 2011.
\newblock Reading digits in natural images with unsupervised feature learning.

\bibitem[{Roy, Jaiswal, and Panda(2019)}]{roy2019towards}
Roy, K.; Jaiswal, A.; and Panda, P. 2019.
\newblock Towards spike-based machine intelligence with neuromorphic computing.
\newblock \emph{Nature}, 575(7784): 607--617.

\bibitem[{Selvaraju et~al.(2017)Selvaraju, Cogswell, Das, Vedantam, Parikh, and
  Batra}]{selvaraju2017grad}
Selvaraju, R.~R.; Cogswell, M.; Das, A.; Vedantam, R.; Parikh, D.; and Batra,
  D. 2017.
\newblock Grad-cam: Visual explanations from deep networks via gradient-based
  localization.
\newblock In \emph{Proceedings of the IEEE international conference on computer
  vision}, 618--626.

\bibitem[{Sengupta et~al.(2019)Sengupta, Ye, Wang, Liu, and
  Roy}]{sengupta2019going}
Sengupta, A.; Ye, Y.; Wang, R.; Liu, C.; and Roy, K. 2019.
\newblock Going deeper in spiking neural networks: Vgg and residual
  architectures.
\newblock \emph{Frontiers in neuroscience}, 13: 95.

\bibitem[{Sharmin et~al.(2020)}]{sharmin2020inherent}
Sharmin, S.; et~al. 2020.
\newblock Inherent Adversarial Robustness of Deep Spiking Neural Networks:
  Effects of Discrete Input Encoding and Non-Linear Activations.
\newblock \emph{arXiv preprint arXiv:2003.10399}.

\bibitem[{Soen and Sun(2021)}]{soen2021variance}
Soen, A.; and Sun, K. 2021.
\newblock On the Variance of the Fisher Information for Deep Learning.
\newblock \emph{Advances in Neural Information Processing Systems}, 34.

\bibitem[{Stigliani, Jeska, and Grill-Spector(2017)}]{stigliani2017encoding}
Stigliani, A.; Jeska, B.; and Grill-Spector, K. 2017.
\newblock Encoding model of temporal processing in human visual cortex.
\newblock \emph{Proceedings of the National Academy of Sciences}, 114(51):
  E11047--E11056.

\bibitem[{Wang, Lin, and Dang(2020)}]{wang2020supervised}
Wang, X.; Lin, X.; and Dang, X. 2020.
\newblock Supervised learning in spiking neural networks: A review of
  algorithms and evaluations.
\newblock \emph{Neural Networks}, 125: 258--280.

\bibitem[{Wu et~al.(2018)Wu, Deng, Li, Zhu, and Shi}]{wu2018spatio}
Wu, Y.; Deng, L.; Li, G.; Zhu, J.; and Shi, L. 2018.
\newblock Spatio-temporal backpropagation for training high-performance spiking
  neural networks.
\newblock \emph{Frontiers in neuroscience}, 12: 331.

\bibitem[{Wu et~al.(2019)Wu, Deng, Li, Zhu, Xie, and Shi}]{wu2019direct}
Wu, Y.; Deng, L.; Li, G.; Zhu, J.; Xie, Y.; and Shi, L. 2019.
\newblock Direct training for spiking neural networks: Faster, larger, better.
\newblock In \emph{Proceedings of the AAAI Conference on Artificial
  Intelligence}, volume~33, 1311--1318.

\bibitem[{Xiao, Rasul, and Vollgraf(2017)}]{xiao2017fashion}
Xiao, H.; Rasul, K.; and Vollgraf, R. 2017.
\newblock Fashion-mnist: a novel image dataset for benchmarking machine
  learning algorithms.
\newblock \emph{arXiv preprint arXiv:1708.07747}.

\bibitem[{Zhang and Li(2020)}]{zhang2020temporal}
Zhang, W.; and Li, P. 2020.
\newblock Temporal spike sequence learning via backpropagation for deep spiking
  neural networks.
\newblock \emph{arXiv preprint arXiv:2002.10085}.

\bibitem[{Zheng et~al.(2020)Zheng, Wu, Deng, Hu, and Li}]{zheng2020going}
Zheng, H.; Wu, Y.; Deng, L.; Hu, Y.; and Li, G. 2020.
\newblock Going Deeper With Directly-Trained Larger Spiking Neural Networks.
\newblock \emph{arXiv preprint arXiv:2011.05280}.

\end{thebibliography}

\end{document}